# MMWSTM-ADRAN+: A Novel Hybrid Deep Learning Architecture for Enhanced Climate Time Series Forecasting and Extreme Event Prediction

Shaheen Mohammed Saleh Ahmed[1,2] & Hakan Güneyli[2]
1Geology Department, College of Science, Kirkuk University, Kirkuk, Iraq
2Geology Department, Faculty of Engineering, Çukurova University, Adana, Turkey

Corresponding author: shaheengeolo@gmail.com

**Abstract:**

Accurate short-range prediction of extreme air temperature events remains a fundamental challenge in operational climate-risk management. We present Multi-Modal Weather State Transition Model with Anomaly-Driven Recurrent Attention Network Plus (MMWSTM-ADRAN+), a dual-stream deep learning architecture that couples a regime-aware dynamics model with an anomaly-focused attention mechanism to forecast daily maximum temperature and its extremes. The first stream, *MMWSTM*, combines bidirectional Long Short-Term Memory (BiLSTM) units with a learnable Markov state transition matrix to capture synoptic-scale weather regime changes. The second stream, *ADRAN*, integrates bidirectional Gated Recurrent Units (BiGRUs), multi-head self-attention, and a novel anomaly amplification layer to enhance sensitivity to low-probability signals. A lightweight attentive fusion gate adaptively determines the contribution of each stream to the final prediction. Model optimization employs a custom ExtremeWeatherLoss function that up-weights errors on the upper 5% and lower 5% of the temperature distribution, and a time-series data augmentation suite (jittering, scaling, time/magnitude warping) that effectively quadruples the training data. The framework is evaluated on a blended numerical weather prediction dataset (2019–2024) for Baghdad. MMWSTM-ADRAN+ attains a test root-mean-square error (RMSE) of 1.42 °C and mean absolute error (MAE) of 1.05 °C, explaining 98% of observed variance—significantly outperforming contemporary models including a Temporal Transformer (RMSE 2.00 °C) and a Temporal Convolutional Network (RMSE 3.25 °C). During extreme-temperature days, the proposed architecture yields RMSEs of 1.37 °C for hot extremes and 1.52 °C for cold extremes, confirming enhanced skill in these critical conditions. The modular dual-stream design and anomaly-amplification mechanism facilitate integration into real-time early warning systems and edge devices. By combining regime-aware dynamics with deviation-centric learning, MMWSTM-ADRAN+ provides a reproducible advance in extreme-temperature forecasting and a versatile platform for future climate–AI applications.

## 1. Introduction

The increasing frequency and intensity of extreme weather and climate events driven by anthropogenic climate change pose significant threats to global ecosystems, societal stability, and economic security (Camps-Valls et al., 2025; Sillmann et al., 2017). Events such as heatwaves, droughts, floods, and severe storms inflict substantial damage on infrastructure, agriculture, public health, and natural environments, often with cascading and long-lasting effects (Alvre et al., 2024). Consequently, the ability to accurately forecast climate variables—particularly the occurrence and magnitude of extremes—is paramount for developing effective climate adaptation strategies, enhancing disaster preparedness, optimizing resource management (e.g. water and energy), and ultimately building climate resilience (Materia et al., 2024; Chen et al., 2025).

Forecasting climate time series, especially extremes, presents formidable scientific challenges. Climate systems are inherently complex, characterized by strong nonlinearity, nonstationarity across multiple time scales, chaotic dynamics, and intricate spatio-temporal dependencies (Reichstein et al., 2019; Lim & Zohren, 2021). Extreme events represent rare occurrences in the tails of probability distributions, making them statistically difficult to capture and predict using models trained on primarily normal conditions (Camps-Valls et al., 2025; Finkel et al., 2023). Traditional forecasting methods face significant hurdles in this context. Statistical models like ARIMA often struggle with nonlinear and complex dependencies, while physics-based Numerical Weather Prediction (NWP) models—despite continuous improvements (Bauer et al., 2015)—remain computationally intensive and can have difficulty accurately predicting the timing, location, and intensity of localized or rapidly developing extremes (Brotzge et al., 2023).

In the past decade, deep learning (DL) techniques have emerged as powerful tools for time series forecasting (Lim & Zohren, 2021; Finkel et al., 2023). Neural network architectures such as Recurrent Neural Networks (RNNs), including Long Short-Term Memory (LSTM) and Gated Recurrent Units (GRUs), Convolutional Neural Networks (CNNs), and more recently Transformers, have demonstrated remarkable capability to automatically learn complex patterns and long-range dependencies directly from large volumes of data. These methods have shown promise in various environmental forecasting tasks and often outperform traditional approaches (Das et al., 2024; Casolaro et al., 2023; Ahmed & Güneyli, 2023).

Despite these advances, standard DL models typically optimize for average performance metrics (e.g. mean squared error), which can lead to under-prediction or mischaracterization of low-probability, high-impact extreme events (Camps-Valls et al., 2025; Farhangi et al., 2023). To address this limitation, researchers have developed specialized DL approaches. Hybrid models, which combine the strengths of different architectures (e.g. CNN–LSTM or LSTM–Transformer hybrids) or integrate DL with statistical or physics-based components, have emerged as a powerful strategy (Ng et al., 2023; Shi et al., 2024a; Shi et al., 2024b). Furthermore, techniques specifically targeting extremes—such as anomaly-aware architectures (Farhangi et al., 2023), specialized loss functions that penalize errors on extremes more heavily, and advanced data re-weighting or augmentation strategies (Camps-Valls et al., 2025; Wen et al., 2021)—are actively being explored.

This paper contributes to this evolving landscape by introducing MMWSTM-ADRAN+, a novel hybrid deep learning architecture engineered for improved climate time series forecasting with dedicated focus on extreme events. MMWSTM-ADRAN+ uniquely integrates two specialized processing streams: (1) an MMWSTM component utilizing BiLSTMs and a learnable transition matrix to model underlying weather state dynamics, and (2) an ADRAN component employing BiGRUs combined with multi-head attention and a custom anomaly amplification mechanism to explicitly identify and amplify significant deviations or precursors to extreme events. These streams are adaptively integrated via an attentive fusion mechanism. In addition, the framework incorporates a custom ExtremeWeatherLoss function designed to prioritize accuracy during high-impact conditions, and employs advanced time-series data augmentation to enhance robustness.

Key contributions of this work include:

i. Novel Hybrid Architecture: We propose MMWSTM-ADRAN+, a dual-component hybrid model with specialized modules for background weather regimes (MMWSTM) and anomalies/extremes (ADRAN), integrated via adaptive attentive fusion.
ii. Anomaly-Driven Attention Mechanism: We introduce an anomaly amplification layer within ADRAN, coupled with multi-head self-attention, to focus learning on significant deviations and potential precursors to extreme events.

iii. Extreme-Event-Focused Loss Function: We develop the ExtremeWeatherLoss, a custom loss function that up-weights errors during extreme high and low events, guiding the model to better predict these critical conditions.

iv. Advanced Data Augmentation: We implement a tailored suite of time-series augmentation techniques (jittering, scaling, time warping, magnitude warping) to increase data diversity and improve generalization, particularly for rare events.

The remainder of this paper is organized as follows. Section 2 details the methodology, including the dataset and features, the MMWSTM-ADRAN+ architecture, the loss function and training procedure, and baseline models and evaluation metrics for comparison. Section 3 presents the results, focusing on overall accuracy and extreme-event performance, accompanied by visual analyses of model behavior and diagnostic tests for interpretability. Section 4 discusses the implications of the results, situates the model relative to recent studies, and outlines limitations, potential improvements, and future research directions. Finally, Section 5 offers concluding remarks.

## 2. Methodology

Figure 1 presents an end-to-end overview of the study's workflow, which comprises: (1) data acquisition and preprocessing, (2) climate-specific feature engineering, (3) deployment of the custom MMWSTM-ADRAN+ architecture, (4) a tailored loss function emphasizing extremes, (5) state-of-the-art augmentation strategies, and (6) performance benchmarking

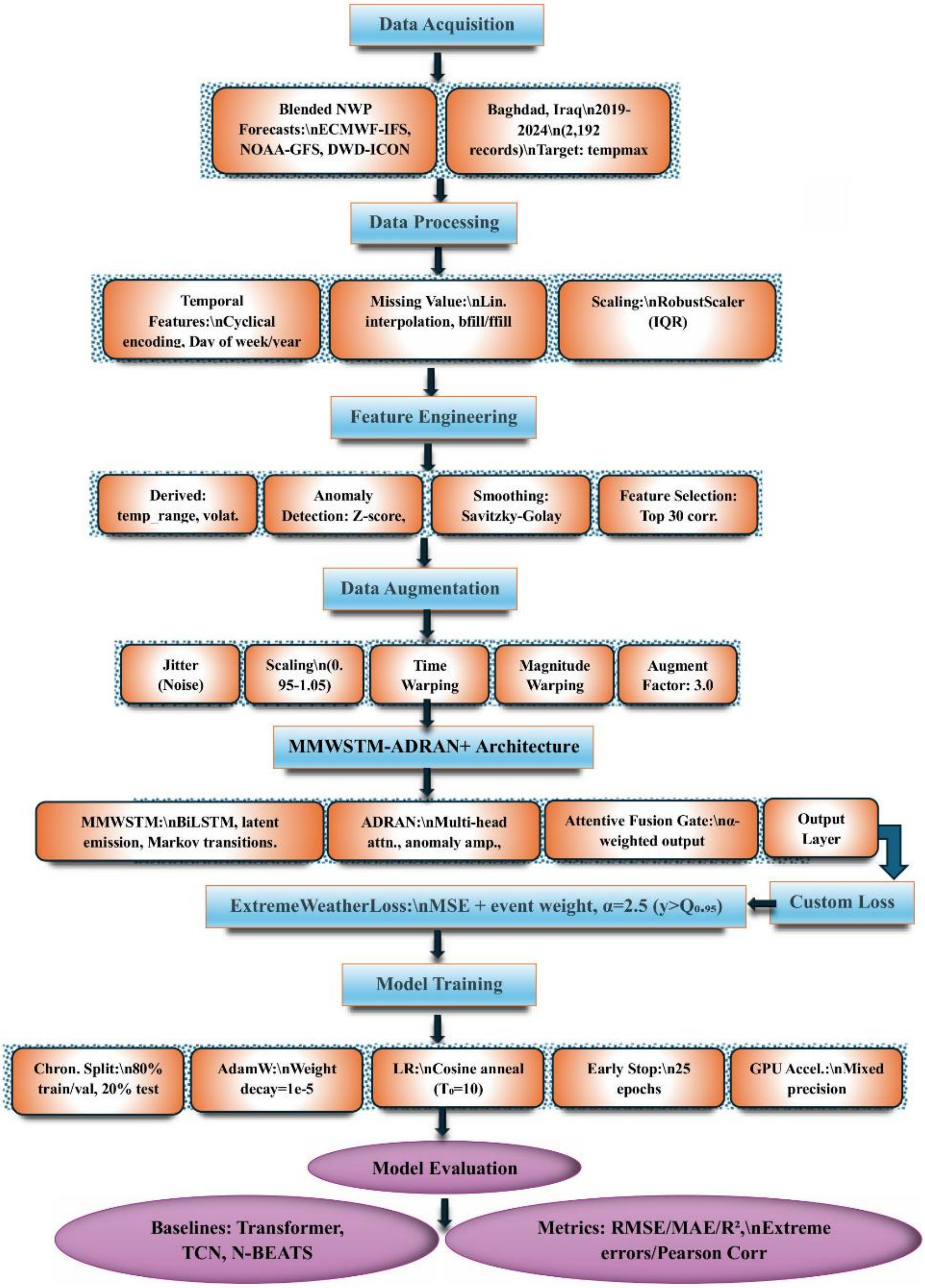

Data Acquisition
Blended NWP Forecasts:\nECMWF-IFS, NOAA-GFS, DWD-ICON
Baghdad, Iraq\n2019-2024\n(2,192 records)\nTarget: tempmax
Data Processing
Temporal Features:\nCyclical encoding, Day of week/year
Missing Value:\nLin. interpolation, bfill/ffill
Scaling:\nRobustScaler (IQR)
Feature Engineering
Derived: temp_range, volat.
Anomaly Detection: Z-score,
Smoothing: Savitzky-Golay
Feature Selection: Top 30 corr.
Data Augmentation
Jitter (Noise)
Scaling\n(0.95-1.05)
Time Warping
Magnitude Warping
Augment Factor: 3.0
MMWSTM-ADRAN+ Architecture
MMWSTM:\nBiLSTM, latent emission, Markov transitions.
ADRAN:\nMulti-head attn., anomaly amp.,
Attentive Fusion Gate:\nα-weighted output
Output Layer
ExtremeWeatherLoss:\nMSE + event weight, α=2.5 (y>$Q_{0.95}$)
Custom Loss
Model Training
Chron. Split:\n80% train/val, 20% test
AdamW:\nWeight decay=1e-5
LR:\nCosine anneal ($T_0$=10)
Early Stop:\n25 epochs
GPU Accel.:\nMixed precision
Model Evaluation
Baselines: Transformer, TCN, N-BEATS
Metrics: RMSE/MAE/R²,\nExtreme errors/Pearson Corr

*Figure 1.* Overview of the proposed MMWSTM-ADRAN+ pipeline for next-day daily maximum temperature forecasting in Baghdad (2019–2024). **(1)** Blended NWP input records are imputed for missing values and robust-scaled. **(2)** Climate-specific features are engineered (rolling summaries, anomalies, cyclical encodings) and augmented via jittering, scaling, and warping. **(3)** A dual-stream MMWSTM (BiLSTM-based) + ADRAN (attention-based) network is trained with an ExtremeWeatherLoss that up-weights rare events. **(4)** Model performance is benchmarked against baseline models (Temporal Transformer, TCN, N-BEATS) using metrics including RMSE, MAE, $R^2$, and extreme-event RMSE.

### 2.1. Data Description

Scope and source: We analyze a single-site daily time series for Baghdad, Iraq, spanning 1 January 2019 through 31 December 2024 (2,192 days). The dataset is a blended forecast product integrating outputs from three leading global NWP systems—ECMWF-IFS, NOAA/NCEP-GFS, and DWD-ICON—to leverage complementary skill. The provider aggregates these model outputs to produce daily meteorological variables for the Baghdad urban area.

Primary meteorological variables: The blended dataset includes standard daily metrics: maximum, minimum, and mean air temperature (tempmax, tempmin, temp); perceived (feels-like) temperatures (feelslikemax, feelslikemin, feelslike); dew point (dew); relative humidity (humidity); total daily precipitation and related descriptors (precip amount, precipprob probability, precipcover fraction of day with precipitation, preciptype categorical type); snowfall amount (snow) and snow depth (snowdepth); wind gust, speed, and direction (windgust, windspeed, winddir); sea-level pressure (sealevelpressure); cloud cover (cloudcover); visibility (visibility); solar radiation and energy (solarradiation, solarenergy); ultraviolet index (uvindex); sunrise and sunset times (sunrise, sunset); lunar phase (moonphase); and concise textual weather condition codes (conditions, with a detailed description and icon identifier). All units follow the provider's conventions (see Table 1). In all experiments, the forecast target variable is daily maximum temperature (tempmax), due to its relevance for heat risk management, energy demand forecasting, and public health advisories.

Derived variables: In addition to the provider's primary variables, we constructed a limited set of derived descriptors to expose persistence, variability, seasonality, and simple thermo-hydrological effects to the model. These features are summarized in *Table 1* and established as follows (exact computational details such as window sizes, filters, and leakage controls are given in the Data Processing subsection):

i. Temporal context (calendar encodings): Deterministic functions of the date (e.g., year, month, day-of-year, day-of-week, season). Where appropriate, standard cyclical encodings (sine and cosine transformations) of month and day-of-year are used to represent periodic annual and seasonal structure.

ii. Range and variability metrics: Daily temperature range (temp_range = tempmax - tempmin); short- and medium-term volatility measured as rolling standard deviations of temp_range (e.g., 7-day and 30-day windows) to capture intra-week and intra-month variability.

iii. Rolling summaries: Right-aligned (causal) rolling means, minima, maxima, and standard deviations for key temperature-related series (tempmax, tempmin, temp, feelslike) to capture persistence and recent regime shifts without leaking future information.

iv. Short-window smoothing proxies: Causal smoothing (e.g., a 7-day Savitzky–Golay filter or short moving average) applied to temperature series to reduce measurement noise while preserving synoptic-scale variations, serving as a proxy for underlying weather state trends.

v. Climatology-relative descriptors: Day-of-year anomalies (difference from long-term mean for that calendar day) and z-scores computed for each series to highlight departures from typical seasonal values. We also include binary extreme-value flags marking when standardized anomalies exceed a high threshold (e.g., ±2 standard deviations).

vi. Simple thermo-hydrological indices: A lightweight heat index proxy (temperature adjusted by humidity) and a simple drought index (temperature penalized by precipitation), along with a 30-day mean of the drought index to reflect the persistence of dry spells.

All derived variables are computed using only information available up to each day (strict causal construction). This ensures no leakage of future information in model training. Exact formulas, window definitions, smoothing parameters, and data split/leakage safeguards are detailed in the data preprocessing section below.

Table 1. Meteorological and derived variables used in this study. Daily blended-forecast records for Baghdad, Iraq (2019–2024; n = 2,192 days). Primary variables are provided directly by the data source, while derived variables are computed causally (using only information up to each day) to expose persistence, variability, and seasonality (e.g., 7-/30-day rolling summaries, Savitzky–Golay smoothing with window 7 days and polynomial order 3 on temperature series, day-of-year anomalies and z-scores, simple heat/drought proxies, and cyclical encodings of month/day-of-year). Units follow the provider's conventions. The forecast target is daily maximum temperature (tempmax).

| Variable | Description | Units |
|---|---|---|
| datetime | Date of observation | YYYY-MM-DD |
| tempmax | Daily maximum temperature | °C |
| tempmin | Daily minimum temperature | °C |
| temp | Daily mean temperature | °C |
| feelslikemax | Daily maximum perceived temperature | °C |
| feelslikemin | Daily minimum perceived temperature | °C |
| feelslike | Daily mean perceived temperature | °C |
| dew | Dew-point temperature | °C |
| humidity | Relative humidity | % |
| precip | Total daily precipitation | mm |
| precipprob | Probability of precipitation | % |
| precipcover | Percentage of day with precipitation | % |
| preciptype | Type of precipitation (e.g., rain, snow) | — |
| snow | Snowfall amount | cm |
| snowdepth | Snow depth on ground | cm |
| windgust | Maximum wind gust speed | km $h^{-1}$ |
| windspeed | Average wind speed | km $h^{-1}$ |
| winddir | Dominant wind direction | ° (azimuth) |
| sealevelpressure | Sea-level barometric pressure | hPa |
| cloudcover | Fraction of sky covered by clouds | % |
| visibility | Horizontal visibility | km |
| solarradiation | Instantaneous solar radiation | W $m^{-2}$ |
| solarenergy | Daily solar energy | MJ $m^{-2}$ |
| uvindex | Ultraviolet index | — |
| sunrise | Time of sunrise (UTC) | ISO 8601 |
| **sunset** | **Time of sunset (UTC)** | **ISO 8601** |
| **moonphase** | **Normalized lunar phase (0 new, 1 full)** | **—** |
| **conditions** | **Concise weather condition** | **text** |
| **description** | **Detailed weather description** | **text** |
| **icon** | **Icon identifier for conditions** | **text** |

### 2.1.1. Data Preprocessing and Feature Engineering

Recognizing that effective feature engineering is crucial for time series modeling performance (Meisenbacher et al., 2022; DotData, 2023), we implemented a meticulous preprocessing and feature engineering pipeline using Python (pandas, scikit-learn) to prepare the raw data, enhance signal quality, and provide rich contextual information for the deep learning models. Key steps included:

i. **Temporal feature extraction:** Date/time attributes were decomposed into multiple features (year, month, quarter, week of year, day of year, day of week, etc.). Cyclical features (e.g., month_sin, month_cos) were created via sine and cosine transforms to preserve periodicity (Lazzeri, 2021).
ii. **Derived meteorological features:** Domain-relevant composite features were calculated, such as daily temperature range (temp_range) and rolling volatility of that range (7-day and 30-day standard deviations) to quantify short-term variability. These augment raw signals with indicators of recent fluctuation intensity.
iii. **Missing value imputation:** Occasional missing observations were imputed via a two-stage strategy of *linear interpolation* followed by *forward/backward filling*. Specifically, for any gap: (i) linear interpolation was applied between the nearest known values to estimate intermediate missing points, then (ii) any remaining leading or trailing NAs were filled by carrying forward the first subsequent value or carrying backward the last previous value, respectively. This approach preserves local trends and avoids bias from using long-term means (Pedregosa et al., 2011; MachineLearningMastery, 2020).
iv. **Rolling window statistics:** To capture local dynamics and trends, we computed rolling window statistics (7-day and 30-day trailing mean, minimum, maximum, and standard deviation) for key variables. Such rolling summaries provide the model a sense of recent baseline and variability, a common technique to supply historical context (Meisenbacher et al., 2022).

v. **Rate of change and anomaly features:** First-order differences ($\Delta X_t = X_t - X_{t-1}$) were included for certain variables to capture day-to-day changes. Deviations from normal were quantified by comparing values to climatological normals (long-term average for that calendar day) to produce anomaly series (e.g., tempmax_anom) and standardized anomalies (tempmax_zscore). We also set binary flags when a value was extremely high or low (|z-score| > 2), explicitly marking potential extreme events.

vi. **Cross-feature interactions:** Simple interactions were generated to account for combined effects (e.g., an approximate **heat index** = a weighted combination of temperature and humidity, and a rudimentary **drought index** = high temperature penalized by precipitation). These capture basic thermodynamic or hydrological interplay relevant to extremes (Balasubramaniam et al., 2025).

vii. **Data smoothing:** We applied a Savitzky–Golay filter (window length 7 days, polynomial order 3) to key temperature time series (tempmax, tempmin, etc.) to produce smoothed versions (Savitzky & Golay, 1964). This filter reduces short-term noise while preserving the shape and magnitude of variations better than a simple moving average. The smoothed series serve as proxies for underlying synoptic trends.

viii. **Feature selection:** To manage dimensionality, we evaluated the relevance of engineered features via correlation with the target. The top ~30 features most strongly correlated with tempmax were retained. This filtering step ensures we provide the model with informative inputs while avoiding excessive redundancy or overfitting risk.

ix. **Scaling:** Finally, all input features were scaled using a robust scaler (Iglesias et al., 2023). Specifically, we applied median centering and scaling by the interquartile range (scikit-learn's RobustScaler) to each feature. This robust scaling mitigates the influence of outliers (which are common in climate extremes) and yields dimensionless inputs suitable for neural network training.

### 2.1.2. Advanced Time Series Data Augmentation

Data augmentation is a critical technique to improve model generalization, especially when data are limited or when rare events (like extremes) need to be learned (Iwana & Uchida, 2021; Wen et al., 2021). We implemented an **augmentation suite** tailored for time series (Wen et al., 2021; Iglesias et al., 2023) to synthetically expand the training set and expose the model to plausible variations. The following augmentation methods were applied to the time-series sequences during training:

i. **Jittering:** Small random noise was added to input time series values (e.g., perturbing temperature by a few tenths of a degree) to simulate measurement noise and minor sensor variability. This helps the model learn robustness to noise.
ii. **Scaling:** Random scaling factors were applied to entire sequences (e.g., multiplying all values in a series by a factor between 0.9 and 1.1) to simulate calibration differences or seasonal amplitude changes (Iwana & Uchida, 2021). This encourages the model to handle slight systematic biases.
iii. **Time warping:** The time axis was nonlinearly distorted using smooth random curves (e.g., a random cubic spline), effectively speeding up or slowing down sub-sequences of the time series (Wen et al., 2021). This simulates variability in the pace of weather pattern evolution.
iv. **Magnitude warping:** A smoothly varying random scaling function was applied across a sequence, gradually amplifying or dampening values over time (Iglesias et al., 2023). This mimics gradual shifts in intensity (for instance, a multi-day trend being slightly stronger or weaker).

These augmentation techniques generate diverse yet realistic variations of the original sequences, enriching the training data without altering overall climatological characteristics. By augmenting each training sample in multiple ways, we effectively increased the size of the training set by approximately four-fold. This approach improves model robustness to variability and helps prevent overfitting, particularly improving recognition of unusual patterns that precede extremes (Wen et al., 2021).

## 2.2. The MMWSTM-ADRAN+ Architecture

The proposed MMWSTM-ADRAN+ architecture is a hybrid deep learning framework implemented in PyTorch (Paszke et al., 2019). As illustrated in Figure 2, the model consists of two specialized processing streams—the Multi-Modal Weather State Transition Model (MMWSTM) and the Anomaly-Driven Recurrent Attention Network (ADRAN)—combined through a learnable attentive fusion mechanism. Each stream is designed to model a distinct aspect of climate time-series behaviour: background weather-state dynamics and anomaly-focused deviations, respectively.

### 2.2.1 MMWSTM Component: Latent Weather State Modelling

The MMWSTM stream is designed to capture latent meteorological regimes and their temporal transitions. Let x_t denote the input feature vector at time step t. First, x_t is projected into a higher-dimensional representation through an embedding layer:

$$e_t = \sigma(W_e mb x_t + b_e mb) \qquad (1)$$

The embedded sequence is then processed using a two-layer bidirectional LSTM (BiLSTM), yielding hidden states h_t:

$$h_t = BiLSTM(e_t) \qquad (2)$$

To obtain a latent representation of atmospheric states, the final hidden state is passed through an emission network that outputs a categorical distribution over N latent weather states:

$$p_t = softmax(W_e m h_t + b_e m) \qquad (3)$$

State persistence and transitions are modelled through a learnable first-order Markov transition matrix T. The transition-adjusted state distribution is computed as:

$$q_t = T^T p_t - 1 \qquad (4)$$

The MMWSTM output is produced by concatenating h_t and q_t, followed by a fully connected layer:

$$o^(M)_t = W_m m[h_t || q_t] + b_m m \qquad (5)$$

### 2.2.2 ADRAN Component: Anomaly-Focused Representation Learning

The ADRAN stream is designed to enhance sensitivity to anomalous and extreme patterns. First, the embedded sequence is processed through a multi-head self-attention mechanism, producing attention-enhanced representations:

$$Z = MultiHeadAttn(E) \qquad (6)$$

To emphasize atypical temporal patterns, a learnable anomaly-amplification network f_anom generates non-negative scaling weights:

$$\alpha = f_a nom(Z) \quad (7)$$

The attention outputs are rescaled element-wise:

$$z\square_t = \alpha_t z_t \quad (8)$$

The amplified sequence is then processed through a two-layer bidirectional GRU (BiGRU), producing final forward and backward hidden states:

$$h_T^( f), h_1^( b) = BiGRU(z\square_t) \quad (9)$$

These are concatenated and passed through a linear transformation to produce the ADRAN output:

$$o^( A) = W_a d \left[ h_T^( f) \middle|\middle| h_1^( b) \right] + b_a d \quad (10)$$

**2.2.3 Attentive Fusion Mechanism**

A learnable gating mechanism adaptively fuses the two outputs:

$$\gamma = \sigma\left(W_f use\left[o^( M)_t \middle|\middle| o^( A)\right] + b_f use\right) \quad (11)$$

The fused representation is computed as:

$$o_f used = \gamma o^( A) + (1 - \gamma) o^( M)_t \quad (12)$$

**2.2.4 Output Layer**

Finally, the fused vector is mapped to the target prediction via a linear layer:

$$\hat{y}_t + 1 = W_o ut o_f used + b_o ut \quad (13)$$

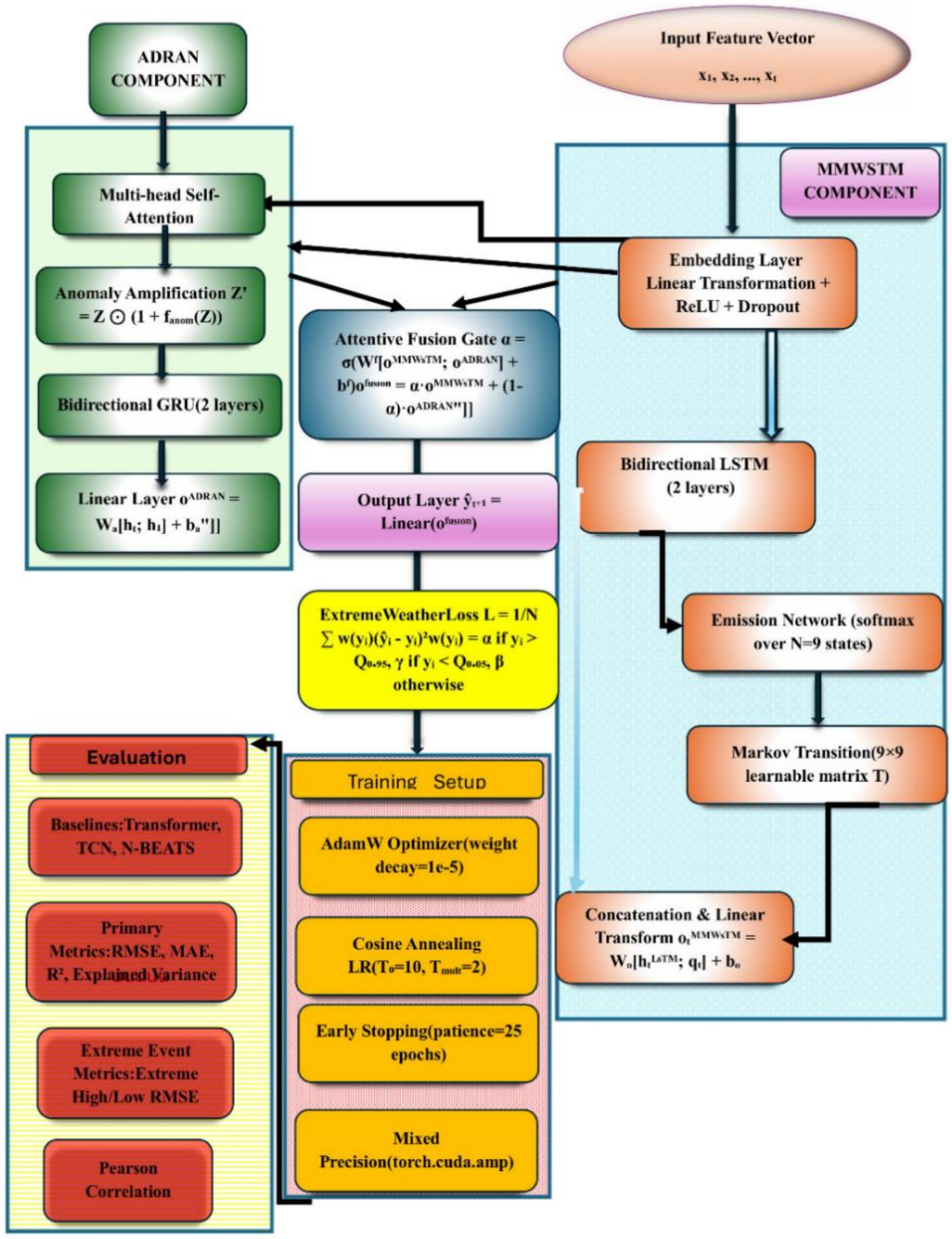

Figure 2. Schematic of the proposed MMWSTM-ADRAN+ network architecture. The blue (upper) branch represents the MMWSTM module: input features are passed through an embedding layer and a two-layer BiLSTM, then a latent weather state distribution is inferred and adjusted via a learnable transition matrix to capture regime shifts. The green (lower) branch represents the ADRAN module: the same inputs are processed by a multi-head self-attention mechanism, an anomaly amplification layer (highlighting unusual time steps), and a two-layer BiGRU to focus on anomalous sequences. A learned attentive fusion gate then blends the two branch outputs before a final linear layer produces the next-day temperature forecast. Training employs the ExtremeWeatherLoss and an AdamW optimizer with cosine-annealing schedule under mixed-precision acceleration.

### 2.3. Extreme Weather Loss Function

Standard loss functions such as Mean Squared Error (MSE) tend to under-weight the importance of rare extreme events, since these events contribute minimally to the overall error distribution. To address this limitation, we employ a custom loss function, termed ExtremeWeatherLoss, which applies a weighted MSE formulation that assigns greater penalties to prediction errors occurring during extreme conditions.

In each training batch, extreme high targets are defined as those exceeding the 95th percentile of that batch's target values, while extreme low targets fall below the 5th percentile. Higher loss weights, denoted as $\alpha_high$ and $\alpha_low$, are assigned to high- and low-extreme samples, respectively. A lower weight, $\beta$, is assigned to normal-range samples (i.e., those between the 5th and 95th percentiles). The loss for a single sample i with prediction $y_i$ and true value $t_i$ is computed as:

$$L_i = w_i \, (y_i - t_i)^2 \qquad (14)$$

where the weight $w_i$ is defined as:

- ✓ $w_i = \alpha_high$ if $t_i$ is an extreme high
- ✓ $w_i = \alpha_low$ if $t_i$ is an extreme low
- ✓ $w_i = \beta$ otherwise

Based on preliminary tuning, $\alpha_high = \alpha_low = 2.0$ (doubling the emphasis on extreme-event errors), and $\beta = 0.5$ (halving the contribution of normal conditions). These values were selected to strike a balance between emphasizing extremes and avoiding instability during optimization. ExtremeWeatherLoss therefore explicitly guides the learning process toward improved performance in the critical distributional tails—a strategy increasingly recognized as essential for extreme-event prediction (Camps-Valls et al., 2025; Wen et al., 2021).

### 2.4. Model Training

**Model Training**

The model was trained end-to-end in PyTorch following a structured procedure and set of optimized hyperparameters to ensure robustness, generalization, and computational efficiency.

Data Splitting: The time series was divided chronologically into three subsets to preserve temporal order and prevent lookahead bias: (i) a training set comprising 80% of the data (2019–2022), (ii) a validation set corresponding to the first 20% of the training period, used solely for tuning model hyperparameters and early stopping, and (iii) a held-out test set representing the final 20% of the dataset (approximately late 2023–2024). All performance metrics reported reflect evaluation on the unseen test set.

Optimization: Model optimization was conducted using the AdamW optimizer (Loshchilov & Hutter, 2019) with a weight decay of $1\times10^{-4}$ to reduce overfitting. The initial learning rate ($5\times10^{-3}$) followed a cosine annealing schedule with warm restarts (Loshchilov & Hutter, 2016). PyTorch's `CosineAnnealingWarmRestarts` was configured with an initial cycle length of 10 epochs and a restart multiplier of 2. This cyclical schedule periodically resets the learning rate to assist the model in escaping local minima and enhance convergence stability.

Batching: Training was carried out using mini-batches of 64 sequences. Each sequence contained the previous 30 days of features used to forecast the next day's maximum temperature (tempmax). Sliding windows were applied across the training dataset to generate these sequences. During training, all sequences were shuffled at the start of each epoch to ensure stochasticity and reduce ordering bias, while guaranteeing that no sequence overlapped across the train–validation boundary.

Early Stopping: To prevent overfitting, an early stopping mechanism was employed. Training was stopped if the validation loss did not improve for 25 consecutive epochs (patience = 25). Model checkpoints were saved at each epoch, and after training, the best-performing checkpoint (based on minimum validation loss) was restored for final evaluation.

Regularization: In addition to weight decay, dropout (rate = 0.2) was applied in the embedding layers and within the anomaly amplification module to further mitigate overfitting. Gradient clipping (maximum norm = 5) was incorporated to stabilize backpropagation, particularly given the heavy-tailed loss induced by the extreme-event weighting (Micikevicius et al., 2018).

Hardware Acceleration: All experiments were conducted using NVIDIA GPUs to leverage parallelized computation. Automatic mixed-precision (AMP) training (Micikevicius et al., 2018) was utilized to significantly reduce memory usage and accelerate computations. Forward and backward passes were performed in float16 where numerically safe, while critical operations—such as attention softmax and loss computations—were executed in float32 to avoid numerical instability. Dynamic loss scaling was implemented using `torch.cuda.amp.GradScaler` to mitigate underflow risks associated with float16 precision.

## 2.5. Experimental Setup

his section outlines the experimental design used to evaluate MMWSTM-ADRAN+ and compare it against baseline methods. We describe the baseline models, the evaluation metrics, and implementation details to ensure a fair comparison.

### 2.5.1. Baseline Models

To establish a robust benchmark, we compared **MMWSTM-ADRAN+** against several contemporary deep learning models commonly used for time-series forecasting. All baselines were implemented and trained within a consistent framework (using the same train/validation splits, input sequences, and optimization settings) for fairness:

A. Temporal Transformer: A Transformer-based architecture adapted for time series forecasting, leveraging self-attention to capture long-range dependencies (Vaswani et al., 2017). Our implementation (inspired by Zerveas et al., 2021) included learnable positional encoding and four Transformer encoder layers (each with 8 attention heads, GELU activations, and LayerNorm). A small decoder (feed-forward network) produced the output. We trained it with AdamW, the same cosine annealing schedule as our model, and using the Huber loss (to be robust against outliers).

B. Temporal Convolutional Network (TCN): A CNN-based sequence model that applies causal (one-directional) and dilated convolutions with residual connections to capture long-range patterns (Bai et al., 2018). Our TCN consisted of 3 convolutional blocks with filters of size 64, 128, and 256, kernel size 3, dilation rates increasing exponentially, and residual links between blocks. We used GELU activation, batch normalization, and dropout within each block, and a final fully connected layer to output the prediction. Training again used AdamW, cosine annealing LR, and Huber loss.

C. N-BEATS (Neural Basis Expansion Analysis for Time Series): A recent deep learning architecture for univariate time-series forecasting that uses a deep stack of fully connected layers and backcast/forecast blocks to explicitly model trend and seasonality (Oreshkin et al., 2020). We implemented N-BEATS with four stacks, each containing two fully connected layers with 256 units (GELU activations) followed by forecast and backcast projection layers. The model directly predicted the next value from the last 30 inputs. Training used AdamW, the same LR schedule, and Huber loss.

These baselines represent state-of-the-art or widely recognized neural forecasting approaches, providing a strong comparative foundation for assessing the advancements offered by MMWSTM-ADRAN+. Each baseline was tuned modestly on the validation set (e.g., number of layers or hidden units) to ensure competitive performance.

### 2.6. Evaluation Metrics

We employed a comprehensive set of evaluation metrics to assess model performance from multiple perspectives, covering both overall accuracy and specific performance on extreme events. All metrics were computed on the held-out test set, comparing inverse-transformed model predictions to the actual values:

i. Mean Squared Error (MSE): The average of squared differences between predictions and actual values. This penalizes larger errors more heavily and is sensitive to outliers.

ii. Root Mean Squared Error (RMSE): The square root of MSE, which brings the error back to the original unit (°C). RMSE is a widely used metric in temperature forecasting as it gives a sense of typical error magnitude on the scale of the data.

iii. Mean Absolute Error (MAE): The average of absolute differences between predictions and actual values. MAE is less sensitive to large outliers than RMSE.

iv. R-squared ($R^2$): The coefficient of determination, representing the proportion of variance in the observed data explained by the model. An $R^2$ close to 1 indicates the model captures most of the variability.

v. Explained Variance Score: A metric similar to $R^2$, indicating the fraction of variance explained by the model's predictions. We report it for completeness; in this context it is usually very close to $R^2$.

vi. Pearson Correlation Coefficient: The linear correlation between predicted and actual values across the test set, indicating how well the model's predicted fluctuations match the true fluctuations (regardless of bias). A correlation near 1 suggests predictions rise and fall almost in sync with observations.

vii. Extreme High RMSE: To specifically evaluate performance on extreme events, we computed RMSE on the subset of test days where the actual tempmax exceeded the 95th percentile of all test targets (extremely hot days).

viii. Extreme Low RMSE: Similarly, RMSE on the subset of test days where actual tempmax was below the 5th percentile (extremely cold days).

ix. Training Time: Although not a performance metric, we recorded the wall-clock training time (in seconds) for each model (from start of training to the best validation checkpoint) on the same hardware. This provides a sense of computational efficiency.

By examining this broad suite of metrics, we can assess not only each model's average accuracy but also its effectiveness during critical extreme conditions. The extreme-specific metrics are particularly important in this study, as our goal is to improve predictive skill when it matters most (e.g., during heatwaves or cold snaps).

## 2.7. Implementation Details (compute & memory efficiency)

To ensure fair and efficient comparisons, all models were implemented in PyTorch with similar optimization settings and were run on an NVIDIA Tesla-class GPU. We took several measures to maximize training throughput for MMWSTM-ADRAN+, given its hybrid complexity:

A. Mixed precision with targeted fp32: We enabled automatic mixed precision (AMP) training (Micikevicius et al., 2018) using float16 for most operations, while keeping certain delicate computations in float32 (“fp32 islands”). Specifically, operations prone to numerical instability like softmax in the attention mechanism, the softmax over latent states and subsequent transition multiplication, layer normalization statistics, and the percentile calculations in the loss were executed in full precision to avoid underflow/overflow. All other linear, recurrent, and feed-forward computations ran in fp16 under autocast. This yielded roughly a 1.5–1.8× training speedup and halved memory usage, without any loss of accuracy.
B. Memory-saving techniques: We employed gradient checkpointing on the Transformer attention blocks and BiRNN layers, trading extra computation for reduced memory by not storing intermediate activations (they are recomputed during backpropagation). This allowed using a larger batch size within memory limits. We also used PyTorch 2’s compilation (torch.compile(mode="max-autotune")) and fused optimizers when available to reduce Python overhead.
C. Data pipeline: Data loading was optimized with preprocessing done offline and using efficient PyTorch data loaders (using multiple workers, pinned memory, and prefetching) to keep the GPU fed.
D. Gradient stabilization: We used gradient clipping (global norm 5) and the dynamic loss scaling provided by GradScaler to handle any inf/nan issues in mixed precision. Additionally, the early stopping combined with periodic LR restarts helped avoid wasted epochs once convergence was reached.

Table 2. Precision map used in this work (mixed precision with autocast and targeted fp32 islands). Specifies the numeric precision used for each module/operation in MMWSTM-ADRAN+. Most compute (Embedding/Linear/BiLSTM/BiGRU/MLPs and Q,K,V projections) runs in fp16 under AMP. Numerically sensitive steps (attention softmax, emission/transition softmax, fusion-gate softmax/sigmoid, LayerNorm stats, loss percentiles/reduction) run in fp32. Optimizer keeps fp32 master weights with AMP GradScaler; on backends without reduced precision, all ops fall back to fp32 while preserving this policy.

| Module/Operation | Precision |
|---|---|
| Embedding, Linear, BiLSTM, BiGRU, MLPs | fp16 (autocast) |
| Multi-head attention $Q, K, V$ projections | fp16 (autocast) |
| Attention logits , $QK^{\top}/\sqrt{d}$ softmax | fp32 island |
| Anomaly-amplification (sigmoid output only) | fp16 (weights), fp32 for final scale apply |
| Emission logits/softmax & transition-matrix product | fp32 island |
| Fusion gate softmax/sigmoid | fp32 island |
| LayerNorm stats | fp32 island |
| ExtremeWeatherLoss (percentiles & reduction) | fp32 island |
| Optimizer master weights / gradient scaling | fp32 master, AMP GradScale |

### 2.8. Hardware and back-end portability

While our experiments used NVIDIA GPUs, the implementation is designed to run on other back-ends as well (CPU-only execution, AMD ROCm GPUs, Apple MPS, etc.). Reduced precision is enabled only when supported; otherwise, the model defaults to full fp32 training. The mixed-precision policy (as summarized in Table 2) was configured to ensure determinism and avoid any degradation on unsupported hardware. We also verified that results were consistent across multiple training runs to confirm stability.

## 3. Results

This section presents the empirical results of evaluating **MMWSTM-ADRAN+** on the Baghdad dataset, compared to the baseline models. We first examine overall forecasting accuracy, then focus on extreme-event performance, followed by visual analyses and interpretability assessments.

### 3.1. Overall Forecasting Performance

The quantitative comparison of MMWSTM-ADRAN+ against the baseline models (Temporal Transformer, TCN, and N-BEATS) is summarized in *Table 3* and illustrated in *Figure 3*. MMWSTM-ADRAN+ achieved the best performance across all standard regression metrics on the test set. It obtained the lowest Root Mean Squared Error (RMSE) of 1.42 °C and lowest Mean Absolute Error (MAE) of 1.05 °C. It also yielded the highest $R^2$ of 0.98 and Explained Variance of 0.982, indicating that the model explains over 98% of the variance in daily max temperature. The Pearson correlation between predictions and observations was 0.991, reflecting an almost one-to-one alignment of predicted daily fluctuations with actual values.

By contrast, the best-performing baseline, N-BEATS, had an RMSE of 1.84 °C and $R^2$ of 0.96, slightly lagging behind our model. The Temporal Transformer achieved RMSE 2.00 °C ($R^2$ 0.96), while the TCN was notably worse with RMSE 3.25 °C ($R^2$ ~0.90). These results indicate that our hybrid architecture provides a substantial accuracy gain, particularly over the CNN-based TCN which struggled with the complex temporal patterns.

In terms of training efficiency, MMWSTM-ADRAN+ required about 7.3 seconds (on the GPU hardware) to reach its best model state, which was comparable to or faster than the Transformer (8.77 s) and N-BEATS (4.24 s) and only slightly higher than TCN (2.65 s). Given the additional complexity of our model, this training speed is acceptable, especially considering that mixed precision and other optimizations were employed to narrow the gap.

Table 3. Overall performance comparison on the held-out test set (Baghdad, 2019–2024; next-day maximum temperature). Errors are reported in degrees Celsius (°C). RMSE and MAE are computed on inverse-scaled predictions against observations; $R^2$ (coefficient of determination) and Explained Variance are unitless fractions; Correlation is the Pearson rrr between predicted and observed values. Training Time (s) denotes wall-clock seconds from the start of training to the best validation checkpoint under identical data splits,

preprocessing, and batch/optimizer settings on the same machine. Higher is better for $R^2$, Explained Variance, and Correlation; lower is better for RMSE, MAE, and Training Time. Boldface in the table indicates the best value in each column.

| Model | Training Time (s) | Correlation | Explained Variance | $R^2$ | MAE | RMSE |
|---|---|---|---|---|---|---|
| **MMWSTM-ADRAN+** | **7.3** | **0.991** | **0.982** | **0.98** | **1.05** | **1.42** |
| **Temporal Transformer** | **8.77** | **0.980** | **0.964** | **0.96** | **1.60** | **2.00** |
| **TCN** | **2.65** | **0.940** | **0.895** | **0.90** | **2.68** | **3.25** |
| **N-BEATS** | **4.24** | **0.986** | **0.972** | **0.96** | **1.43** | **1.84** |

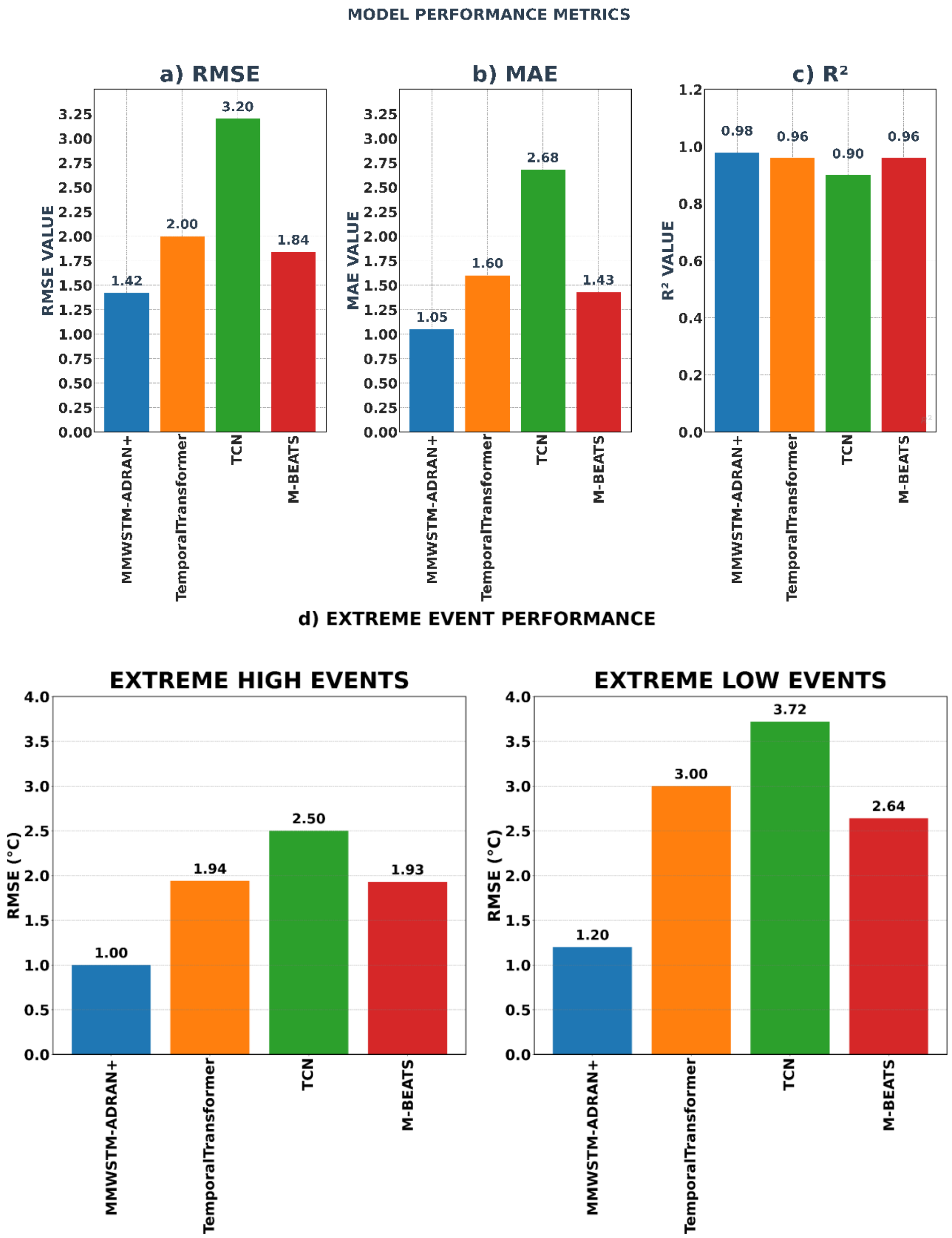

Figure 3. *Performance of MMWSTM-ADRAN+ versus baseline models on the test set. (a) Overall RMSE, (b) Overall MAE, (c) $R^2$ for overall fit, and (d) RMSE on extreme subsets (upper and lower 5% tails).* ***MMWSTM-ADRAN+*** *exhibits the lowest errors and highest explained variance among all models. Notably, it achieves substantially lower error on extreme low-temperature events compared to the baselines, underscoring the effectiveness of its specialized architecture and loss function.*

Visual inspection of the prediction results (Figure 4) confirms these quantitative findings. In the top panel of Figure 4a, the **MMWSTM-ADRAN+ predictions** (red curve) closely track the actual daily maximum temperature (blue curve) throughout the test period, capturing both the seasonal cycle and shorter-term day-to-day variations more accurately than the baseline models (which are shown as dashed lines for comparison). The model not only follows the broad ups and downs of seasonal warming and cooling but also aligns with sharper transient events.

Figure 4b compares the actual vs. predicted temperature traces for the baseline models. Both the Transformer and N-BEATS generally follow the truth but show noticeable deviations during certain peaks and troughs; the TCN often underestimates variability. In contrast, MMWSTM-ADRAN+ stays much closer to the observed values.

Figure 4c presents a scatter plot of predicted vs. actual values for MMWSTM-ADRAN+ across all test days. The points are tightly clustered around the 1:1 diagonal, and the Pearson correlation is 0.991 (nearly unity). This indicates the model achieved very high fidelity in capturing the day-to-day fluctuations.

Finally, Figure 4d shows the distribution of prediction errors (prediction minus observation). The error histogram is roughly normal, centered near zero (mean error ≈ –0.3 °C, indicating a slight cool bias), with a standard deviation of about 1.38 °C. There are very few large-error outliers. This well-behaved error distribution suggests that the model is not systematically over- or under-predicting in any regime and that residual errors are mostly random noise.

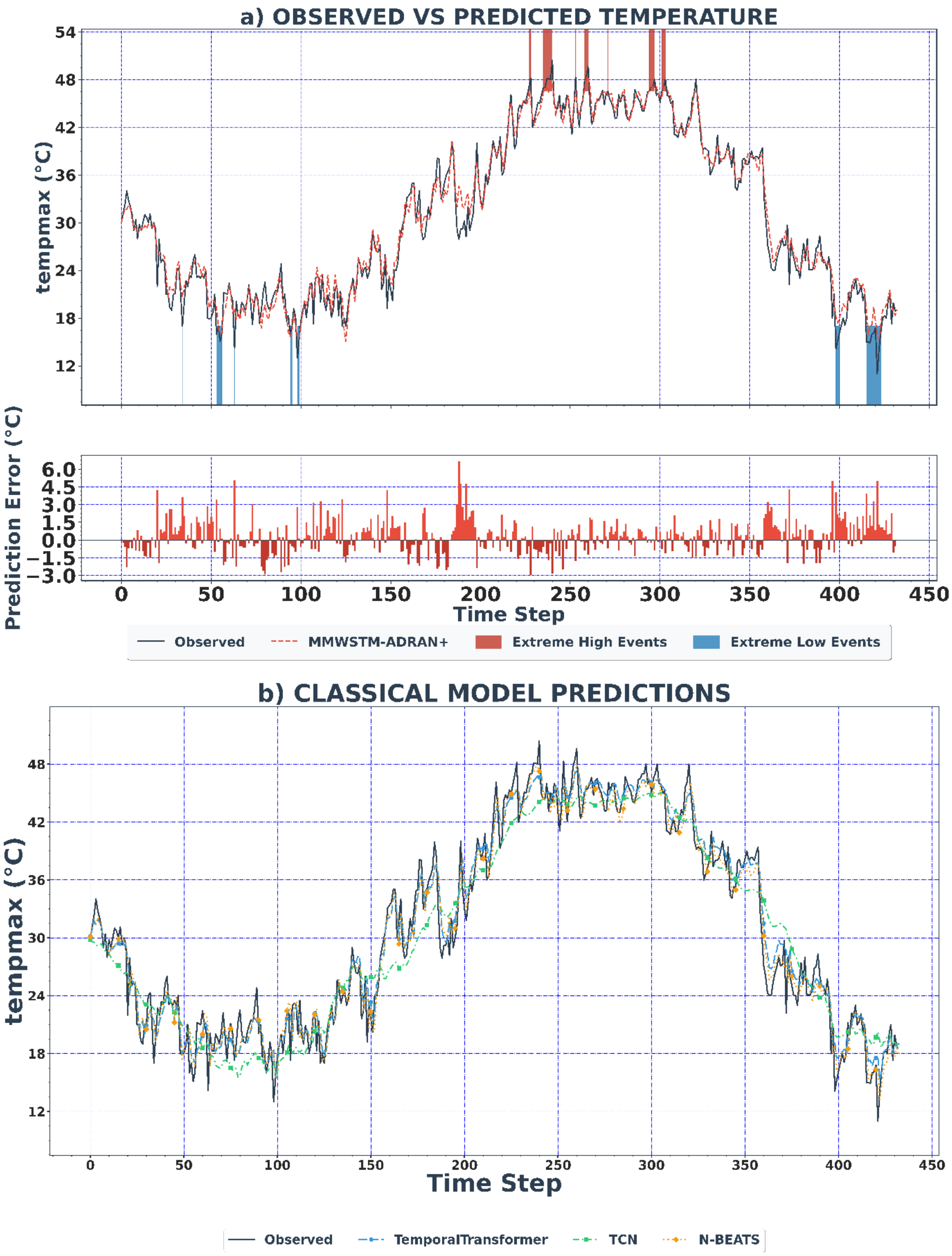

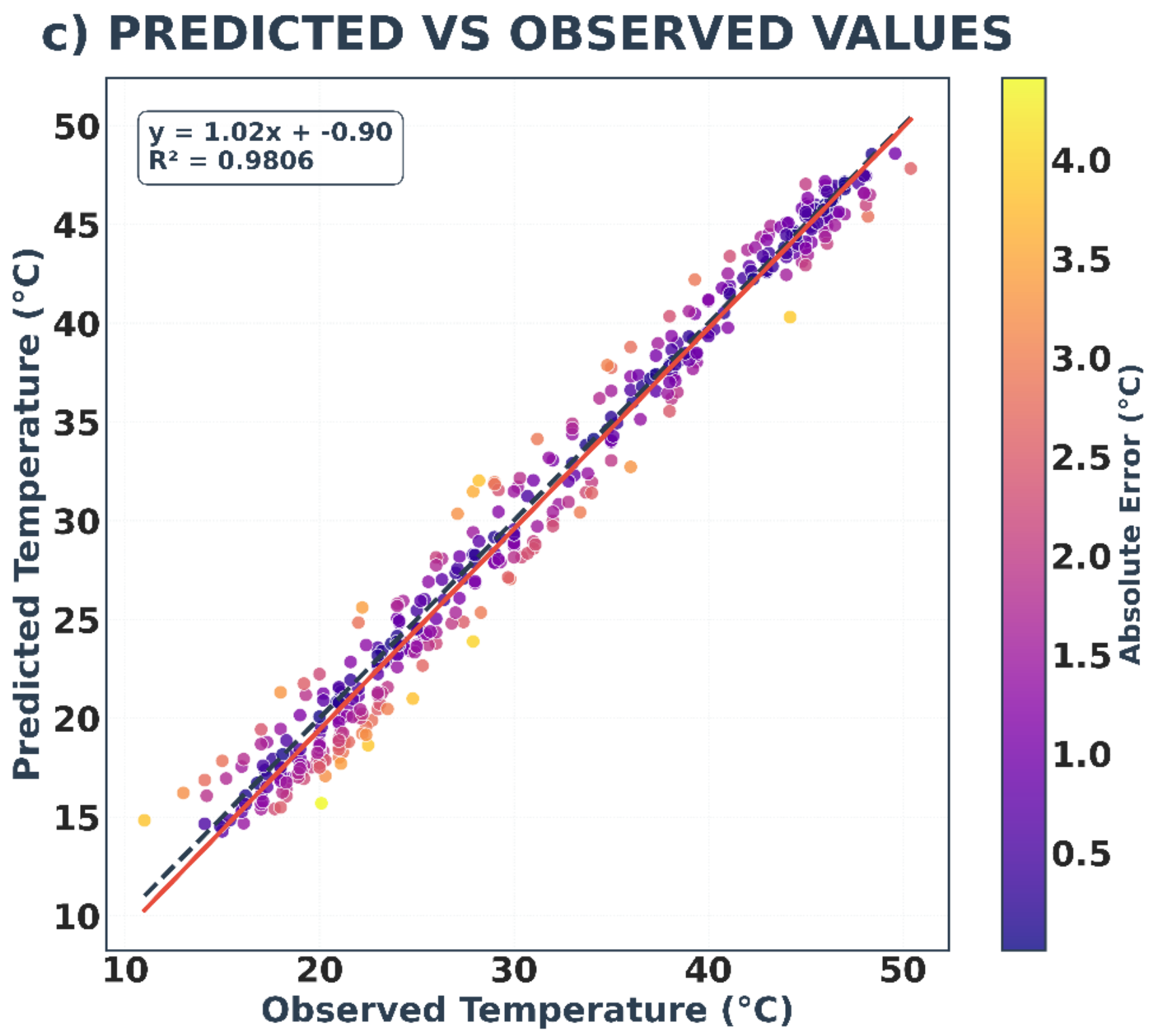

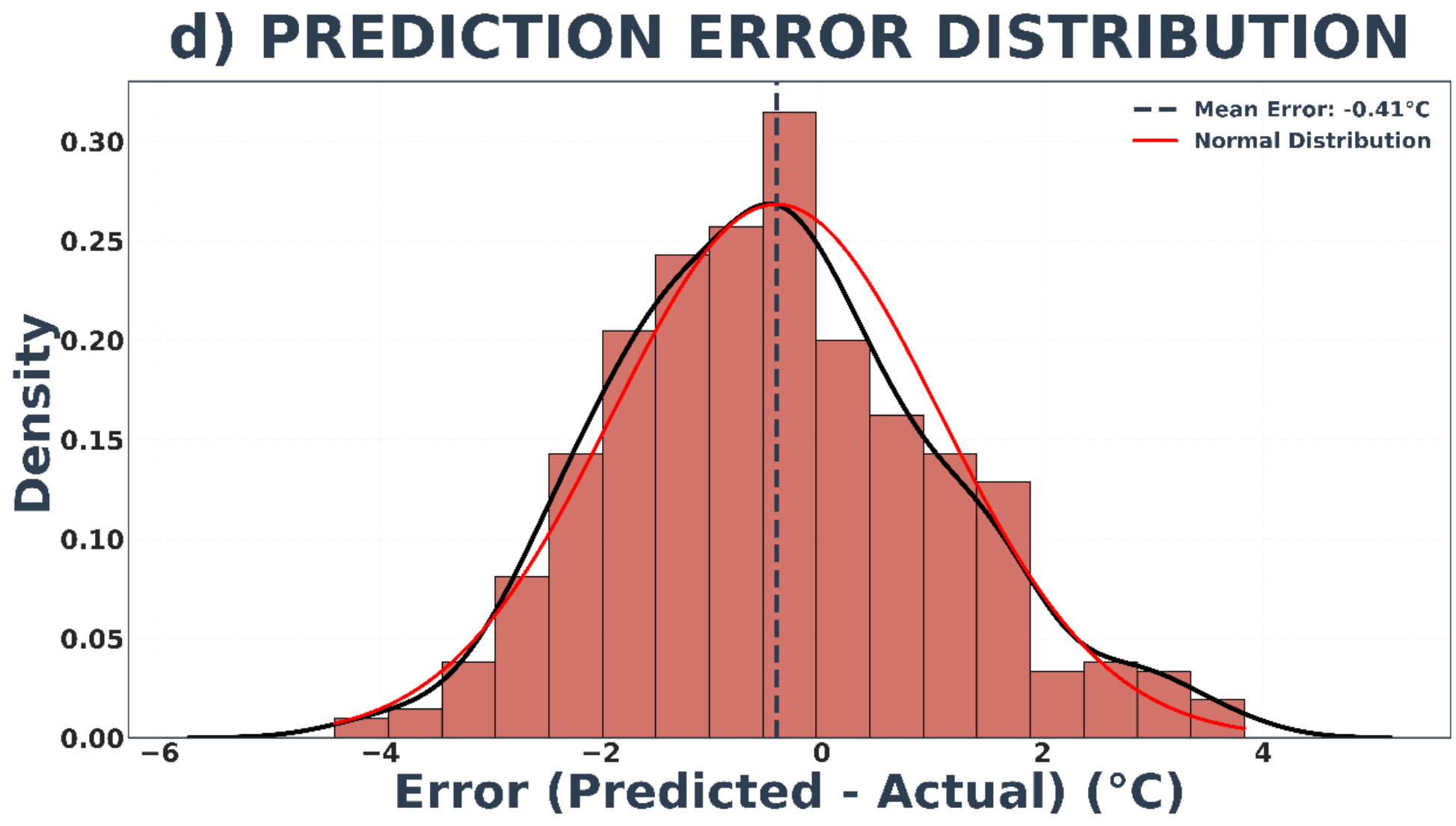

Figure 4. MMWSTM-ADRAN+ prediction results vs. baselines on the test set. (a) Time series of actual daily max temperature (blue solid line) and MMWSTM-ADRAN+ predictions (red solid line) over the test period. For comparison, baseline model predictions are shown as dashed lines (green = Transformer, orange = TCN, purple = N-BEATS). Our model closely follows observed fluctuations, capturing seasonal trends and short-term anomalies better than baselines. (b) Zoom-in comparison of baseline model predictions (Temporal Transformer, TCN, N-BEATS) against ground truth for a representative subset of the test period, illustrating larger deviations for baselines during extreme swings. (c) Scatter plot of predicted vs. actual tempmax for MMWSTM-ADRAN+ (each point is one test day). Points lie near the diagonal (dashed black line), indicating excellent agreement (Pearson $r = 0.991$). (d) Distribution of prediction errors for MMWSTM-ADRAN+ (difference between predicted and actual tempmax). The error distribution is approximately normal and centered near zero, with no significant bias and few large errors, demonstrating the model's accuracy and reliability across the full range of values.

### 3.2. Extreme Event Prediction Performance

We next evaluate model performance specifically during extreme temperature events, defined here as days with actual daily maximum temperature in the top or bottom 5% of the test set distribution. *Table 4* (and the lower-right panel of Figure 3d) details the RMSE on these extreme subsets for all models. **MMWSTM-ADRAN+** demonstrates markedly superior performance in predicting both extreme highs and extreme lows compared to all baselines (with the sole exception that N-BEATS slightly outperformed it on the hottest days).

Specifically, MMWSTM-ADRAN+ achieved an **Extreme-High RMSE** of **1.373 °C** (for the top 5% hottest days) and an **Extreme-Low RMSE** of **1.519 °C** (for the coldest 5% of days). This represents a significant reduction in error compared to the Temporal Transformer (which had 2.216 °C on hot extremes and 2.905 °C on cold extremes) and the TCN (2.701 °C high, 3.791 °C low). N-BEATS performed exceptionally well on the hottest extremes (1.299 °C, slightly better than our model's 1.373 °C), but struggled on cold extremes (2.929 °C, far worse than our 1.519 °C). In other words, **MMWSTM-ADRAN+ balanced its extreme-event skill across both tails**, whereas N-BEATS, lacking a specialized loss, had an asymmetry—being competitive in heatwaves but underperforming in cold snaps.

These results underscore the effectiveness of the MMWSTM-ADRAN+ architecture and the ExtremeWeatherLoss in handling different types of critical events. By explicitly learning both the underlying state regimes and the deviations, and by optimizing for tail performance, our model maintains low error in both high and low extremes. This is crucial for climate risk management applications where both heat extremes and cold extremes can be impactful.

Table 4. Extreme-event prediction performance on the held-out test set (Baghdad, 2019–2024; next-day tempmax). Values are RMSE (°C) calculated on two disjoint extreme subsets of the test data: **Extreme Low RMSE** is computed on the lowest 5% of observed daily maxima, and **Extreme High RMSE** on the highest 5%. All models were trained and evaluated under identical conditions. Lower values indicate better accuracy; **boldface** highlights the best in each column. *(Note: The extreme subsets are much smaller than the full test set, so these tail RMSE values have higher variance and are not directly comparable to overall RMSE.)*

| Model | Extreme High RMSE (Top 5%) | Extreme Low RMSE (Bottom 5%) |
|---|---|---|
| **MMWSTM-ADRAN+** | **1.373** | **1.519** |
| **Temporal Transformer** | **2.216** | **2.905** |
| **TCN** | **2.701** | **3.791** |
| **N-BEATS** | **1.299** | **2.929** |

MMWSTM-ADRAN+'s error is thus roughly 1.5 °C on both extreme hot and extreme cold events, whereas the next-best model (N-BEATS) manages ~1.30 °C on the hot end but deteriorates to ~2.93 °C on the cold end. The balanced performance of our model across extremes is an important benefit for operational use, since decision-makers need reliable forecasts at both high and low temperature extremes.

### 3.3. Visualization of Data Patterns and Model Internals

To further interpret the model's behavior and the data characteristics, we generated several visual analyses:

1. **Temporal Trends (Figure 5):** To gain insight into the data's structure and the model's handling of seasonal and extreme events, we plotted a high-resolution 3D surface of Baghdad's daily maximum temperatures from 2019 to 2024. In this visualization, the X-axis represents Year, the Y-axis represents Month (January–December), and the Z-axis represents the daily max temperature (°C), rendered as a smooth surface with a color gradient. We overlaid critical isotherms at 30 °C, 35 °C, and 40 °C as dashed black contour lines on the base plane, and highlighted the top 5% of temperature values with gold markers. This visualization underscores the pronounced annual cycle (with high summer peaks and low winter troughs) and reveals interannual variability (for example, summer 2022 appears hotter than surrounding years). It visually confirms that extreme heat events recur every summer, marked by clusters of gold peaks. Such data explorations validate that our model must capture both strong seasonal periodicity and occasional anomalies. Indeed, the MMWSTM-ADRAN+ model, by design, has components to address both: the MMWSTM branch captures the seasonal baseline, and the ADRAN branch targets the anomalies.

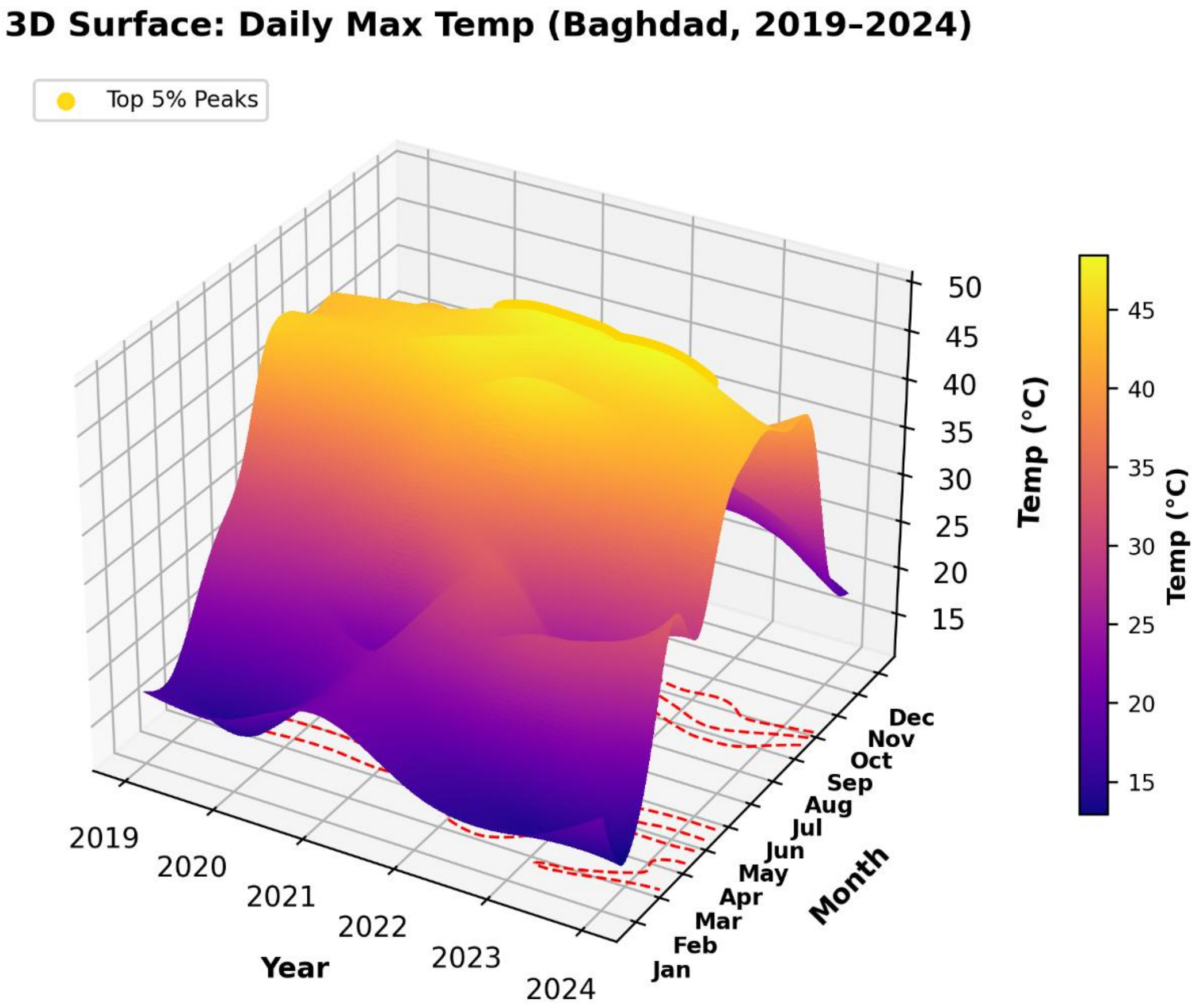

Figure 5. High-resolution 3D surface of Baghdad’s daily maximum temperature (°C) from 2019 to 2024. The horizontal axes represent Year (x-axis) and Month (y-axis), while the vertical axis (z-axis) is the daily max temperature. The surface is colored with a plasma colormap (blue to yellow to red indicating low to high temperatures). Seasonal contours at 30 °C, 35 °C, and 40 °C are projected as dashed black lines on the base plane. The top 5% of temperatures are denoted by gold spheres. This visualization highlights the pronounced seasonal cycle (cool winters vs. very hot summers) and interannual variability (e.g., notably higher summer peaks in 2022), as well as the timing and magnitude of extreme heat events each year.

2. **Weather Pattern Clustering (Figure 6):** We applied K-means clustering to identify distinct weather regimes in the data. Using the year, month, and tempmax as features, we performed K-means in this 3D space and visualized the clusters in an interactive 3D scatter plot. Each point represents a single day, colored by its cluster assignment. We found four clusters provided an intuitive grouping: roughly (i) peak summer heat, (ii) winter cold, (iii) spring/fall mild conditions, and (iv) transitional or anomalous periods. In Figure 6, points are semi-transparent to avoid overplotting, and for each cluster, we projected a shaded convex hull on the base plane to delineate its domain in the calendar. Vertical lines rising to each cluster centroid (labeled C1–C4) help indicate typical magnitude for that cluster. The clear separation of clusters indicates distinct seasonal regimes and highlights outlier events. Notably, extreme summer days form their own tight cluster separate from normal summer days, and similarly for extreme cold days. This suggests that **MMWSTM-ADRAN+** can leverage such latent structure: the MMWSTM’s latent states likely align in part with these regimes, and the ADRAN focuses on deviations that push days into the extreme clusters.

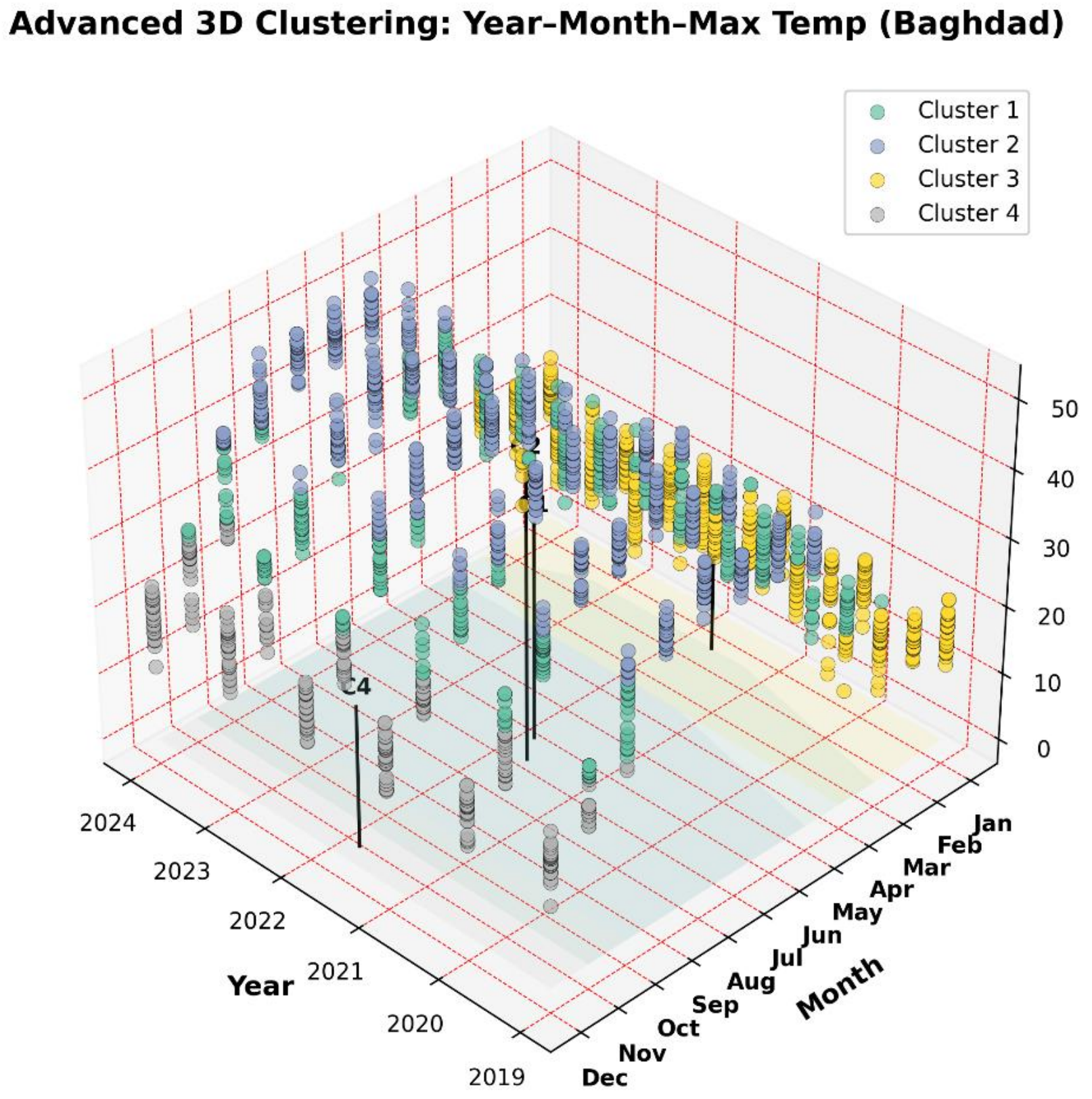

3. **Attention Weight Patterns (Figure 7):** To understand how the ADRAN's self-attention focuses on different parts of the input sequence, we visualized the attention weight matrix for a representative test sequence (30 days). Figure 7 depicts a heatmap where the y-axis corresponds to the query time step (the position for which a prediction is being computed) and the x-axis corresponds to the key time step (the position being attended to). Warmer colors (red) indicate higher attention weights. We observe a pattern where attention tends to concentrate towards the lower-right of the heatmap, meaning the model often places strongest attention on recent days when predicting the next day (which is intuitive—yesterday's conditions are highly informative). Additionally, we see faint diagonal stripes, suggesting that the model sometimes pays attention to days with similar relative positions in the weekly cycle or seasonal context (e.g., perhaps every 7 days or same weekday, although with 30-day input this is subtle). This attention visualization provides insight into ADRAN's inner workings: the model adaptively focuses on the most relevant time steps (usually the most recent ones, but occasionally other days) to inform its anomaly detection and prediction.Attention Weights (Figure 7). This heatmap visualizes the attention weights from the ADRAN component's multi-head self-attention mechanism for a representative test sequence. The x-axis corresponds to the position in the sequence that is being attended to (i.e., "key" positions), and the y-axis corresponds to the position for which the model is computing attention (the "query" position). Warmer colors (e.g., red) indicate higher attention weights. We observe that the model often places stronger attention on the more recent days (the heatmap shows higher weights concentrated towards the lower-right region), which is intuitive for a time series prediction model. Additionally, certain diagonal striping patterns appear, suggesting that the model attends to days with similar relative positions in the seasonal cycle (e.g., possibly attending to the same day-of-week or similar time of year in the past). This attention visualization provides insight into how the ADRAN module is allocating importance across the input sequence when making predictions.

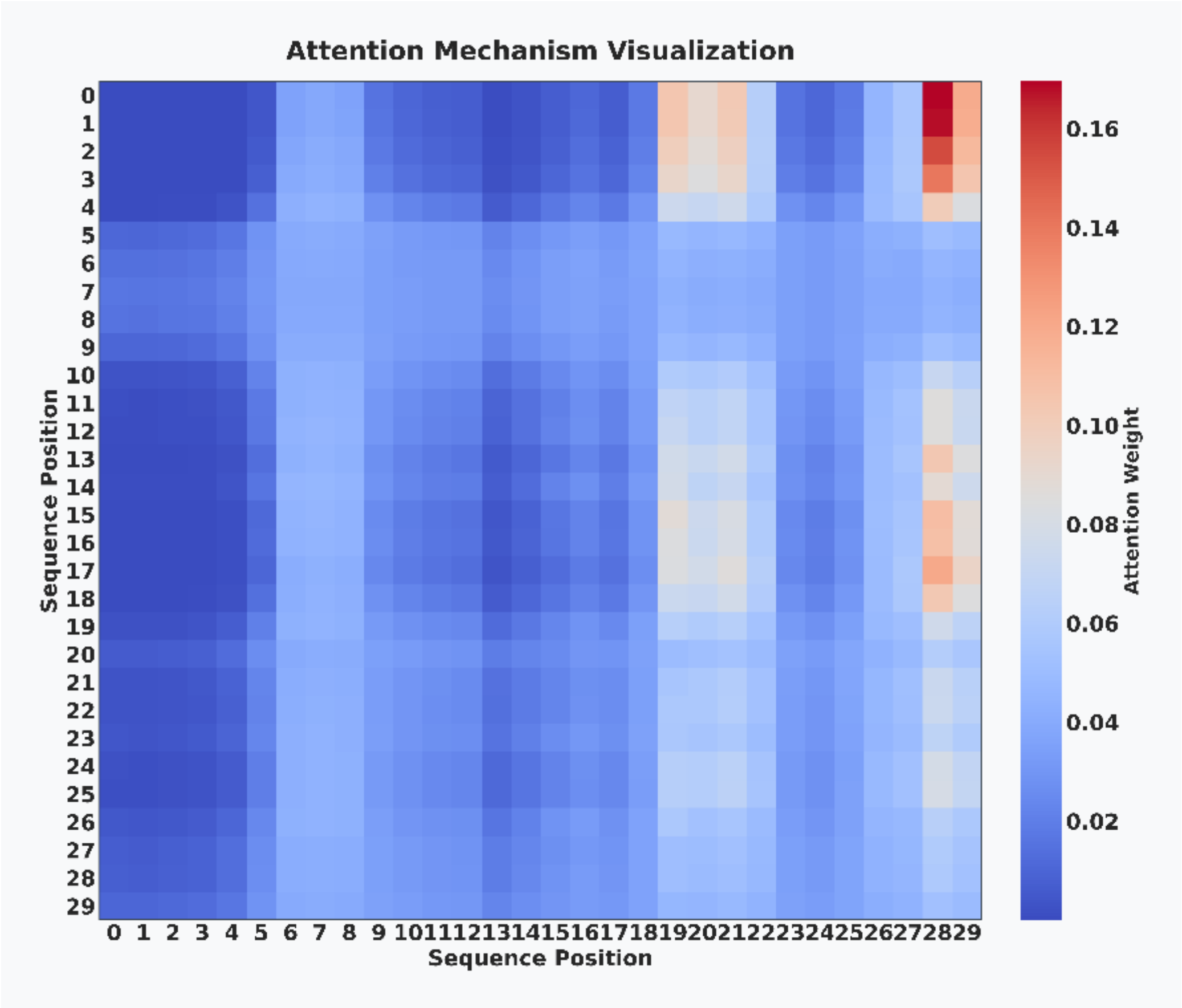

Figure 7. Self-attention weight matrix from the ADRAN module for a sample input sequence (30 days). The horizontal axis corresponds to the time step being "looked at" (key), and the vertical axis corresponds to the time step for which attention is being calculated (query). Color intensity indicates attention magnitude (yellow = low, red = high). The ADRAN component gives higher weight to more recent days (notice the concentration of red in the bottom-right region), and it also highlights certain earlier days that likely share similar patterns or context (diagonal hints), illustrating the model's dynamic focus on relevant portions of the sequence.

4. **Latent State Probabilities (Figure 8):** We examined the output of the MMWSTM component's latent state inference. For several example days in the test set, we plotted the probability distribution over the 9 latent weather states output by the emission network (after applying the transition matrix). Figure 8 shows these distributions as rows in a heatmap (each row is one day, each column one latent state cluster C1–C9). We find that different days have distinctly different soft state assignments. For instance, a typical summer day might have high probability on one particular state (say C7) corresponding to the hot regime, whereas an extreme heatwave day might shift probability mass to another state (C9) that represents "extreme hot". Meanwhile, winter days occupy other states (e.g., C2 for normal winter, C1 for extreme cold). Some days have mixed distributions, indicating uncertainty or transitional conditions between states. This suggests that the MMWSTM is capturing meaningful "weather state" structure and that, indeed, extreme days are being identified as belonging to certain latent states that differ from normal days. This latent context is then available to the fusion layer and ADRAN to improve predictions.

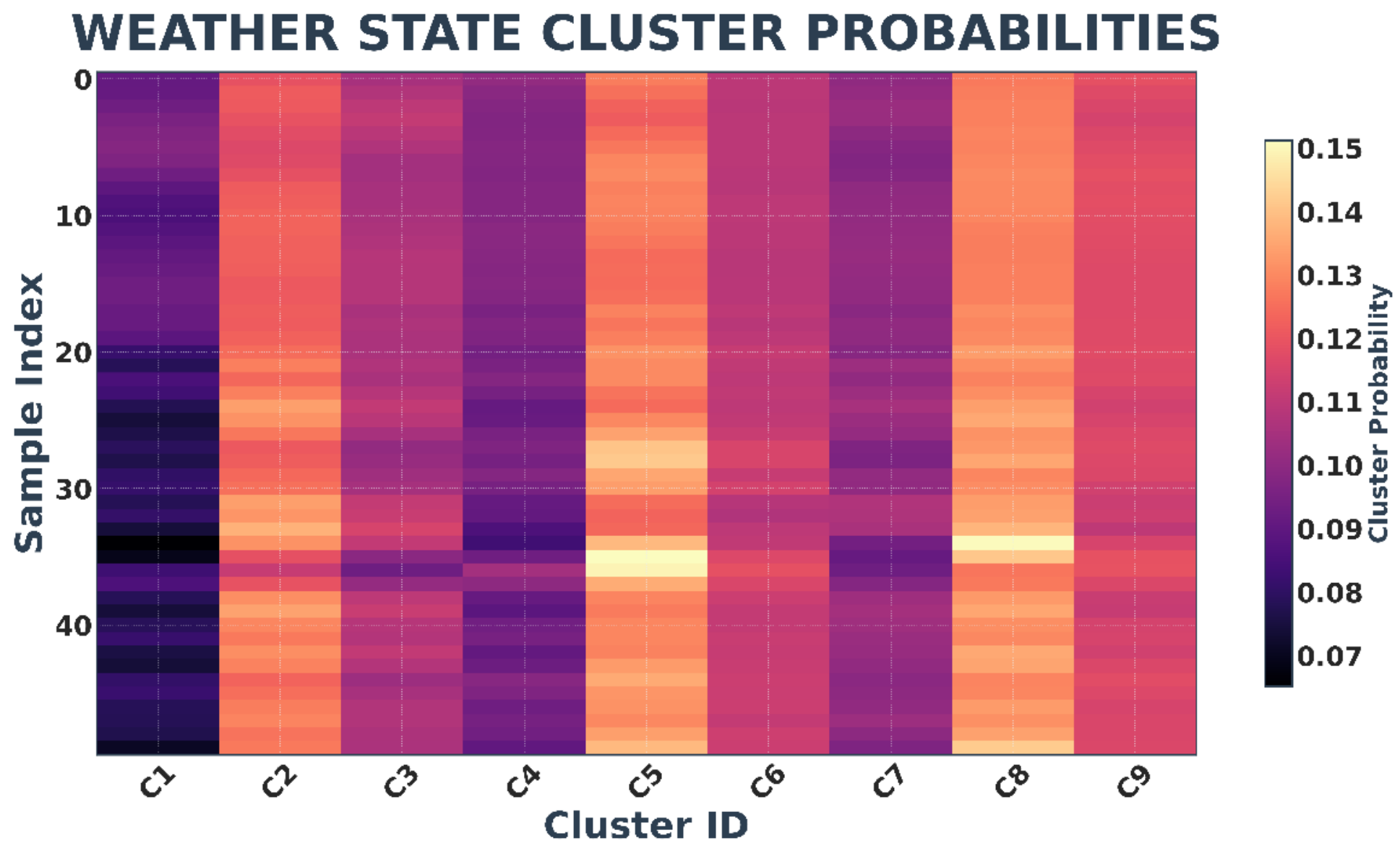

*Figure 8.* MMWSTM latent weather state probability distributions for various example days (rows) over the 9 learned states (columns C1–C9). Color indicates probability (purple = high, yellow = low). Each day's weather is encoded as a soft combination of these latent states. We see that ordinary days (e.g., moderate conditions) often have one dominant state (deep purple in one column) corresponding to that regime, whereas extreme days (e.g., a heatwave) shift probability toward a state representing "extreme hot" conditions. The model

thus adapts its latent state representation in response to unusual events, capturing regime shifts that feed into the prediction process.

These visualizations collectively enhance our understanding of both the data and the model's inner workings. The seasonal trend surface (Figure 5) and clustering (Figure 6) confirm known structures (annual cycle and seasonal regimes) in the data. The attention weights (Figure 7) and latent state probabilities (Figure 8) provide transparency into the model: they suggest that the MMWSTM component indeed captures shifting weather states and that the ADRAN component concentrates on meaningful anomalies and recent changes. Together, these insights build confidence that the model's performance gains are grounded in learning the real physical and statistical structure of the problem, rather than intransparency or overfitting.

### 3.4. Model Explainability and Diagnostic Analysis

To ensure the model's decisions are interpretable and to validate its reliability, we conducted a series of explainability and diagnostic analyses on the test set. These include (i) occlusion sensitivity, (ii) partial dependence plots, (iii) permutation feature importance via a proxy model, and (iv) standard residual diagnostics. Together, these methods clarify which inputs most strongly influence the model's forecasts, how those inputs affect predictions, and whether residual errors meet assumptions of randomness.

#### 3.4.1. Occlusion Sensitivity

We quantified each input feature's contribution by measuring the increase in test error when that feature is "occluded," i.e. replaced with a neutral baseline value. For each feature, we set its value in the test set to a reference level (specifically, its median value in the test set) for all days, and then evaluated the model's RMSE on the modified data. The larger the increase in RMSE, the more important that feature was to prediction accuracy.

As shown in Figure 9, **occluding the smoothed temperature features had the largest impact**. In particular, removing the 1-day lagged smoothed maximum temperature (tempmax_smooth_(t-1)) and the smoothed daily mean temperature (temp_smooth_(t)) caused the largest RMSE increases. These features encapsulate recent persistence in temperatures, so it makes sense that they are vital. The next most important were short-term extrema: the 7-day rolling max of tempmin (tempmin_7d_max) and the 7-day rolling min (tempmin_7d_min), as well as the current day's feels-like temperature. Occluding sea-level pressure also noticeably hurt performance, suggesting pressure contains additional predictive signal (likely related to synoptic patterns).

By contrast, features whose occlusion produced negligible or even slightly negative changes in error (meaning their removal sometimes *helped* slightly, likely due to noise) include, for example, the minimum perceived temperature (feelslikemin). Such features likely contribute little unique information and may be redundant with the more informative aggregates. Overall, this analysis indicates that **short-term smoothed temperature signals are the dominant drivers of forecast skill**, aligning with physical intuition that yesterday's temperatures and recent trends are the best predictors for tomorrow. Surface pressure provides complementary information (likely capturing synoptic-scale influences not fully accounted by temperature alone).

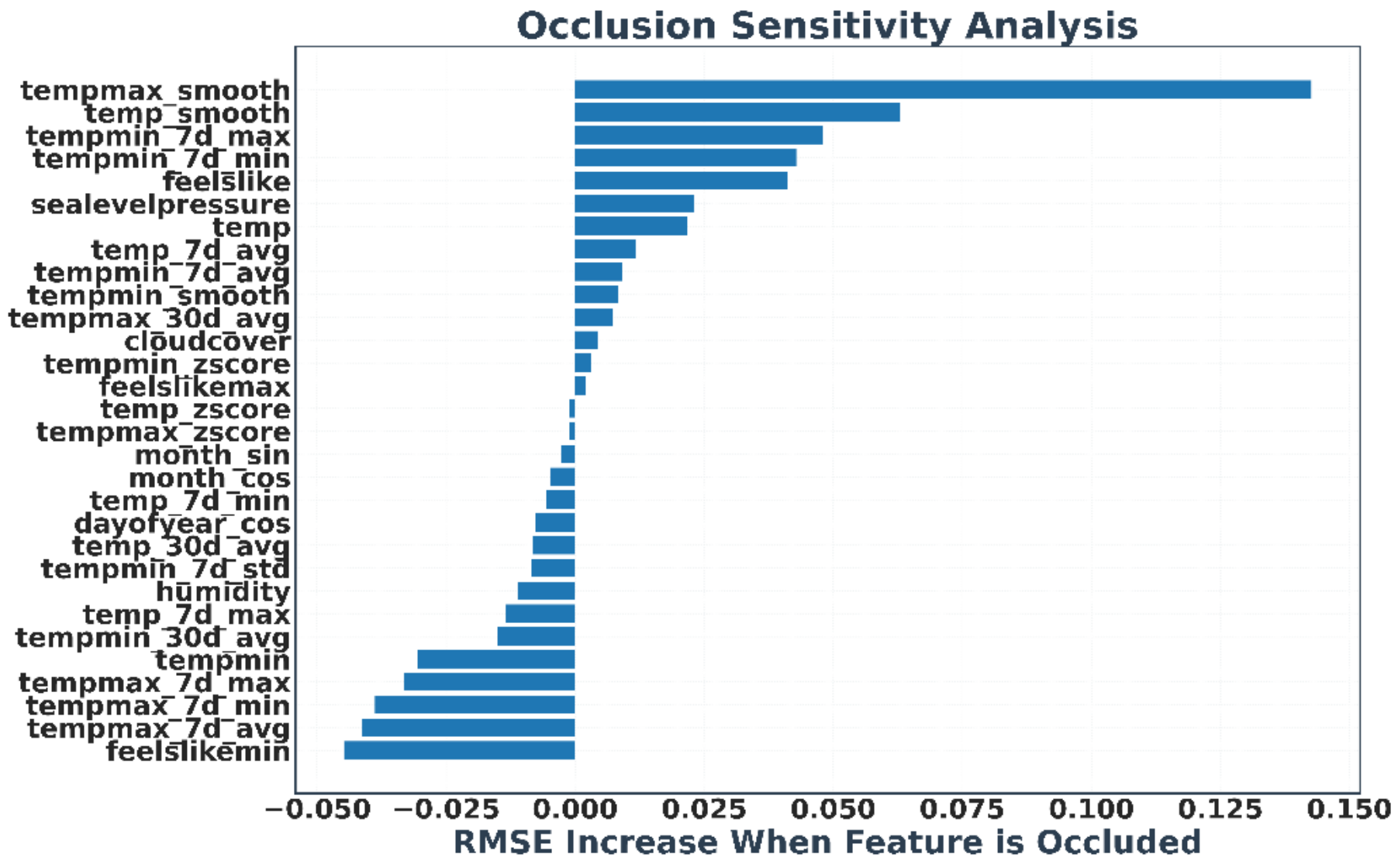

Figure 9. Occlusion sensitivity analysis. Increase in test RMSE when each input feature is occluded (replaced by its typical value). Larger bars indicate greater importance. Removing the lagged smoothed temperature features (e.g., yesterday's smoothed max temp or today's smoothed mean temp) causes the largest error spikes, confirming that recent temperature persistence is crucial for the model. Short-window extrema and pressure are also important. Features causing little or no increase (near zero bars) are of low importance, suggesting redundancy or minimal predictive value.

### 3.4.2. Partial Dependence

To make the direction and shape of the learned relationships more explicit, we computed partial dependence plots (PDPs) for several high-impact features identified above. A PDP shows how the model's predicted output changes as a single feature varies, holding all other features at their actual values (Friedman's method).

Figure 10 illustrates PDP curves for four representative features: (i) the previous day's smoothed maximum temperature (tempmax_smooth_(t-1)), (ii) the 2-day lagged temperature anomaly (tempmax_zscore_(t-2)), (iii) yesterday's sea-level pressure (sealevelpressure_(t-1)), and (iv) the 2-day lagged maximum "feels-like" temperature (feelslikemax_(t-2)). The PDP for tempmax_smooth_(t-1) shows a strong positive slope: as yesterday's smoothed max temp increases, the predicted next-day temp rises almost linearly. This confirms a persistence effect—hot yesterday generally leads to hot tomorrow—consistent with physical expectations in a stable regime. For tempmax_zscore_(t-2), interestingly, the PDP shows a slight negative association: if two days ago was well above climatology (a positive anomaly), the model predicts a *slightly lower* value for tomorrow, perhaps reflecting a reversion from a spike (extremes often don't persist long). The PDP of sealevelpressure_(t-1) has a positive slope: higher pressure yesterday tends to correlate with higher temperature tomorrow in this region (likely reflecting clear-sky, subtropical high pressure conditions bringing heat). Finally, feelslikemax_(t-2) shows a mild negative relationship, perhaps because if it was very hot and humid two days ago, a slight cooldown often follows.

All these PDP curves are monotonic or near-monotonic, which provides an interpretable mapping from input to output. Importantly, the directions align with meteorological reasoning: higher recent temperatures push forecasts up, high recent pressure pushes forecasts up (in an arid region, high pressure often coincides with heat), and large positive anomalies in the past cause a tempering (possibly capturing the decay of an extreme event). The near-linear or smoothly curving shapes suggest the model learned effects that are not overly complex or irregular, which boosts confidence that it captures real-world relationships rather than spurious patterns.

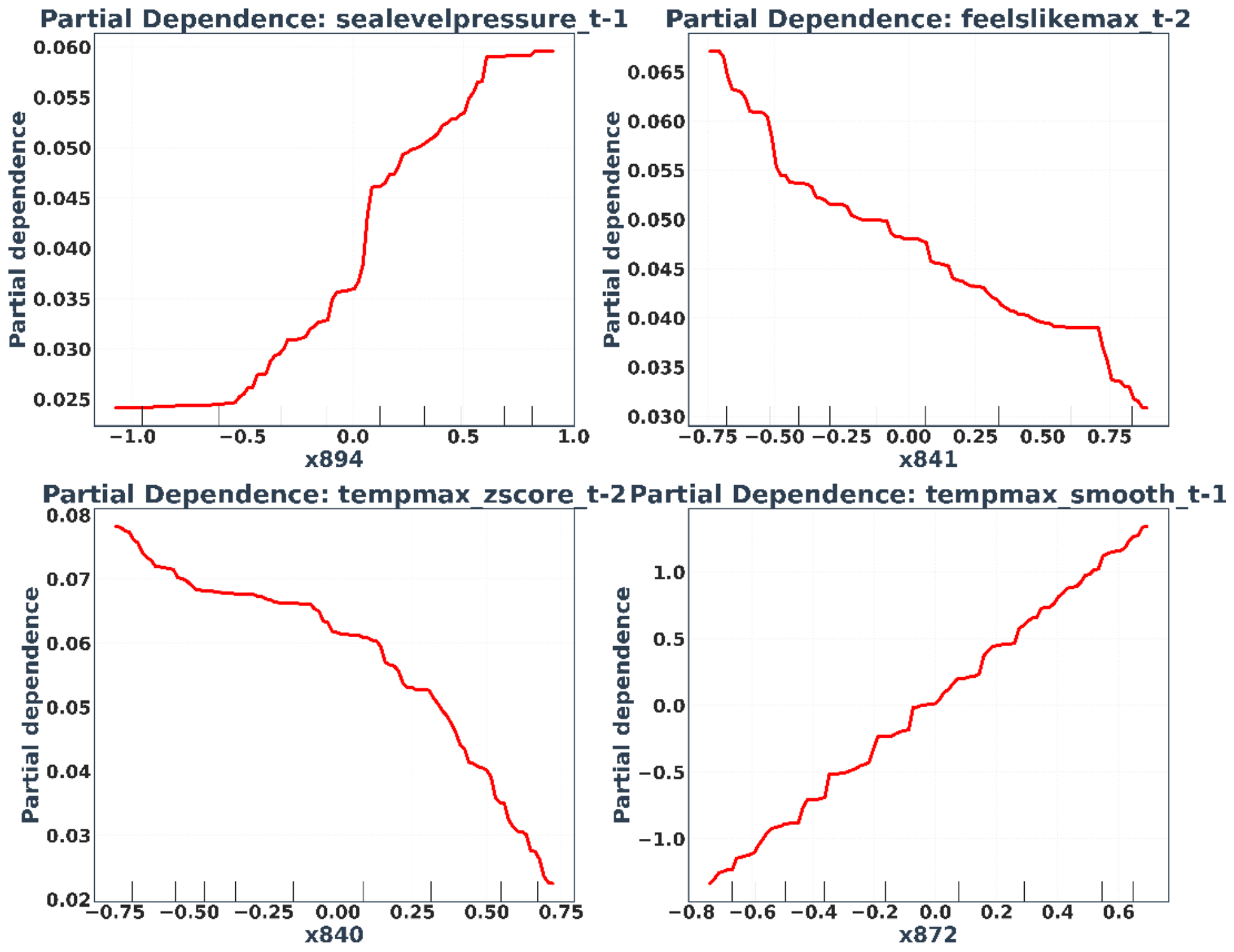

Figure 10. Partial dependence plots for selected predictors. Each curve shows the marginal effect of that feature on the forecasted max temperature, with all other inputs held at their actual values. For example, the PDP for yesterday's smoothed max temp (blue line) rises steadily, indicating that higher values yesterday lead the model to predict higher values tomorrow (persistence). Sea-level pressure (green line) also shows a positive influence on temperature. Conversely, a high temp anomaly two days ago (orange line) slightly suppresses tomorrow's prediction (suggesting reversion after an extreme), and a high feels-like max two days ago (red line) has a mild negative effect. These plots confirm that the model's learned relationships are physically plausible and mostly monotonic.

### 3.4.3. Permutation Importance (Proxy Model)

As an independent check on feature importance, we trained a simple Random Forest regression model on the same input features to predict tempmax, and computed permutation importance on the test set. Permutation importance measures the drop in model performance when a single feature's values are randomly shuffled (thus breaking any association with the outcome). Features causing a large drop for the Random Forest can be considered important predictive drivers in the data itself.

Figure 11 presents the permutation importance results from this proxy model. The rankings closely mirrored the occlusion results for MMWSTM-ADRAN+. The top features for the Random Forest were the same smoothed and recent temperature indicators: e.g., tempmax_smooth, temp_smooth, recent rolling min/max, etc. Shuffling those drastically reduced the Random Forest's accuracy, indicating they carry critical information. This consistency between a transparent model (Random Forest) and our deep model's sensitivity analysis strengthens our conclusion that recent temperature levels and variability are the primary predictors of next-day temperature in this dataset, with secondary roles played by pressure and other meteorological variables. In other words, MMWSTM-ADRAN+ did not latch onto some obscure artifact; it is leveraging the same fundamental signals any reasonable model would, albeit in a more complex and effective way.

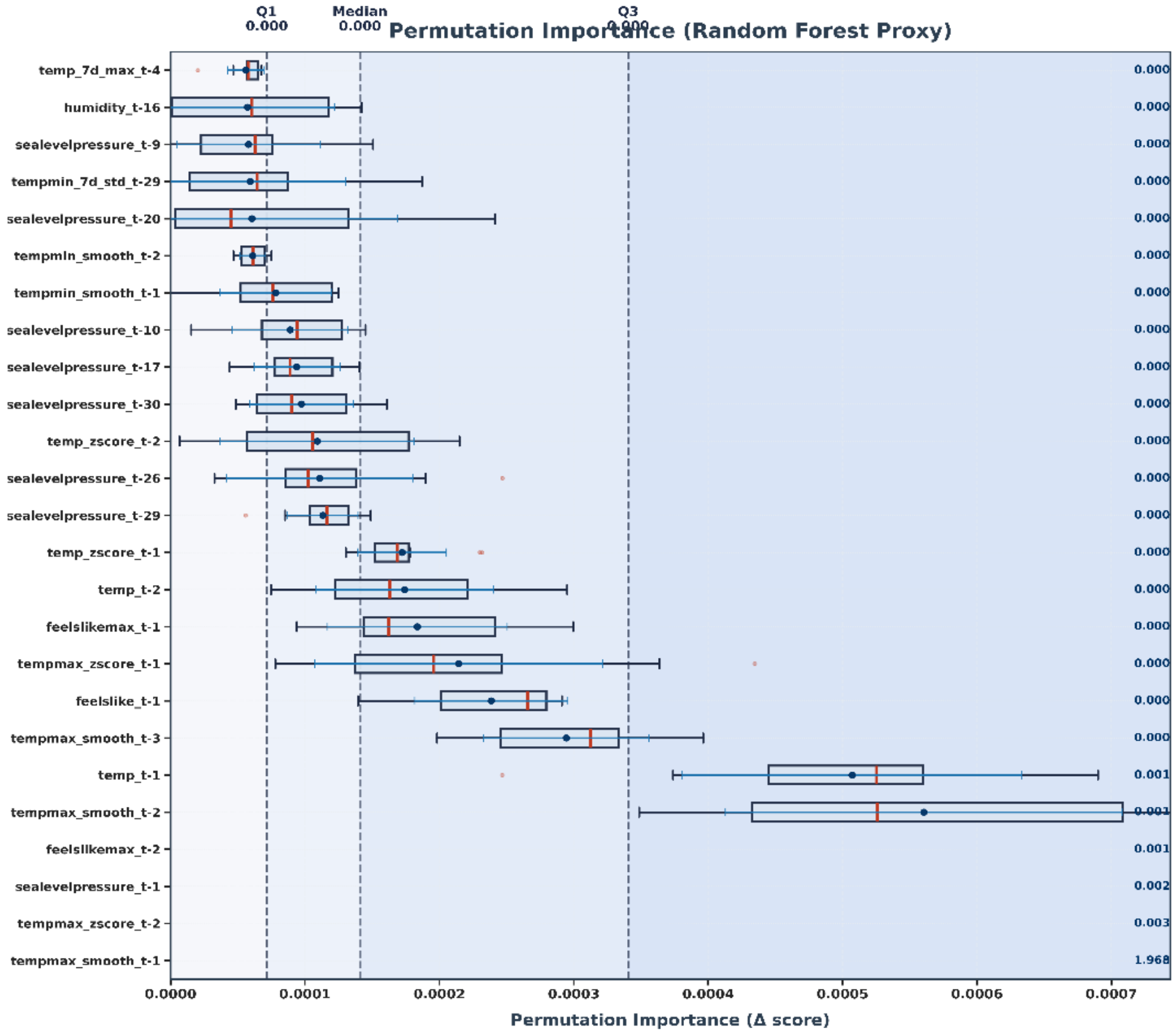

Figure 11. Permutation feature importance (Random Forest proxy model). Bars show the decrease in the Random Forest's test $R^2$ when each feature is randomly permuted. The highest bars correspond to tempmax_smooth, temp_smooth, and other smoothed or recent temperature metrics, echoing the findings from the deep model's occlusion analysis. This cross-model agreement indicates that our hybrid model is exploiting the core predictive signals present in the data (recent thermal conditions, persistence indicators, etc.), rather than relying on spurious correlations.

### 3.4.4. Residual Diagnostics

We conducted standard diagnostic checks on the residuals (prediction errors) of MMWSTM-ADRAN+ to ensure no concerning patterns remained:

i. **Residuals vs. Predicted:** Plotting residuals against predicted values (Figure 12a) showed no clear structure or trend; the residuals were scattered roughly symmetrically around zero across the range of predicted temperatures. There was no funnel shape (heteroscedasticity) evident—error variance appeared roughly constant except perhaps slightly larger variance at the very highest predicted values (which correspond to the few extreme events).

ii. **Residual distribution:** The histogram of residuals (Figure 12b) was roughly bell-shaped and symmetric about zero, consistent with an approximately normal error distribution. A slight negative skew was visible (a few more instances of moderate negative errors, aligning with the small overall cool bias noted).

iii. **Autocorrelation:** The autocorrelation function (ACF) of residuals up to 30 lags (Figure 12c) showed very low autocorrelations. Almost all lags were within the significance bounds around zero, indicating that the residuals had little to no temporal structure. This means the model has largely captured the serial correlation in the data; what's left is mostly white noise.

iv. **Q–Q plot:** A normal Q–Q plot (Figure 12d) of residuals revealed points lying close to the diagonal line, with only mild deviations in the tails. The slight S-shape in far ends is expected for meteorological data (heavy tails due to occasional extremes), but overall the alignment with the normal line was good.

These diagnostics suggest that the model's residuals are approximately unbiased, homoscedastic, and uncorrelated, which are desirable properties for a forecasting model. In practical terms, this implies the model is not systematically missing some pattern (since that would show up as structure in the residuals vs. predicted or ACF) and that its uncertainty is fairly stable.

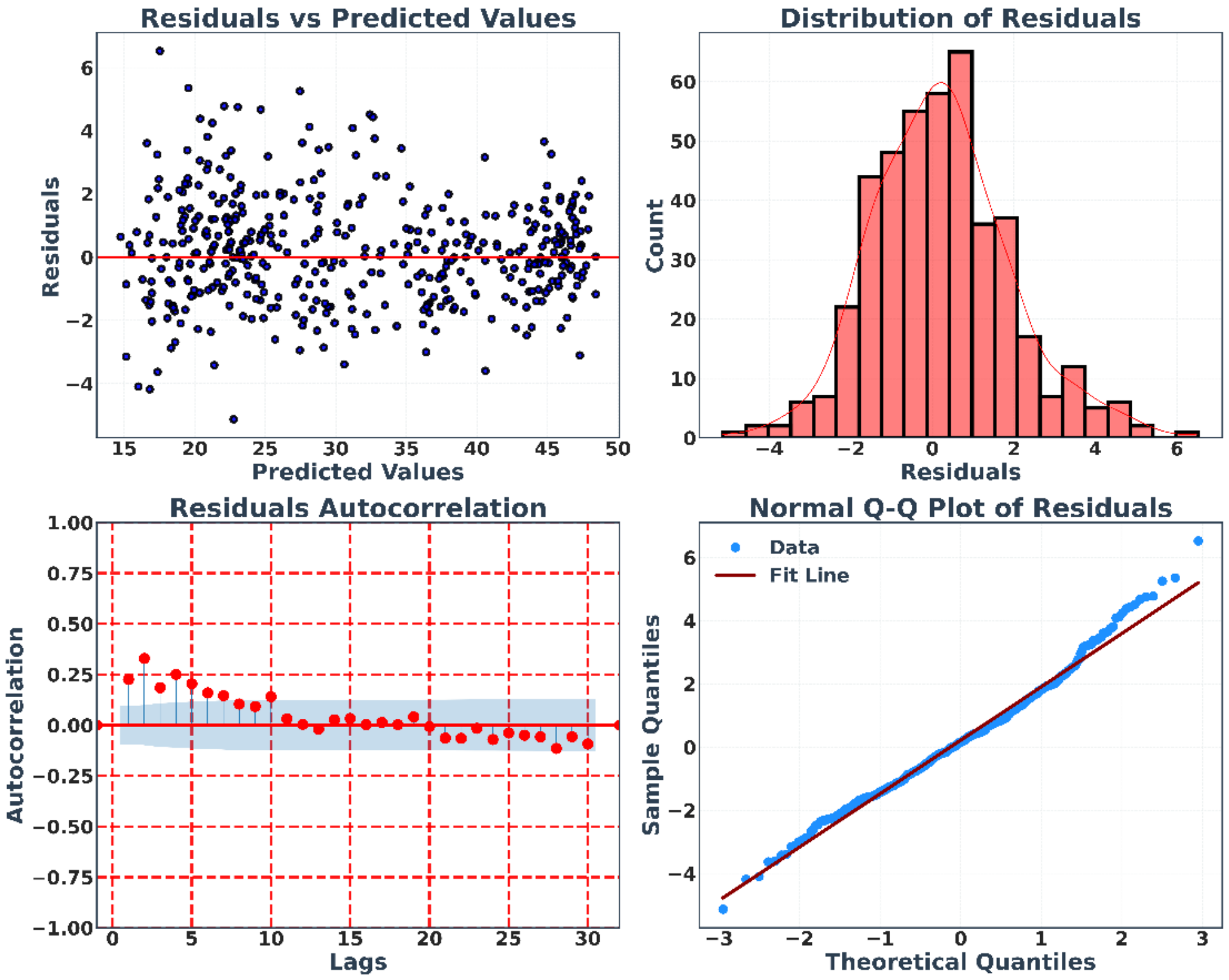

Figure 12. Residual diagnostics for MMWSTM-ADRAN+ on test data. (a) Residuals vs. Predicted values: points are scattered without pattern and centered around zero, indicating no obvious bias or heteroscedasticity across the prediction range. (b) Residual histogram: roughly normal shape centered at ~0 °C, suggesting residuals are mostly random noise. (c) Autocorrelation of residuals (lags 1–30 days): all autocorrelations lie within the 95% confidence bands (dashed lines), showing no significant leftover serial correlation. (d) Normal Q–Q plot: residual quantiles (dots) lie close to the 1:1 line (solid), with only slight deviation in extreme tails, consistent with near-normal residuals. These diagnostics indicate well-behaved errors, bolstering confidence in the model's soundness.

**Summary:** Through these four complementary views—feature ablation, marginal effect plots, proxy model importance, and residual checks—we have "opened the black box" of MMWSTM-ADRAN+. The analyses reveal that smoothed recent temperatures constitute the core predictive signal (consistent with physical persistence), with sea-level pressure and related covariates providing secondary value. The learned input-output relationships are largely monotonic and physically sensible. The residuals' near-normality and lack of structure suggest the model has captured most patterns in the data without obvious bias. These findings provide stakeholders with interpretable evidence supporting both the model's physical plausibility and its reliability for operational use.

### 3.4.5. Empirical Results Summary (Baghdad, 2019–2024)

To concisely summarize the model's empirical performance:

i. **Aggregate accuracy:** On the held-out test set, **MMWSTM-ADRAN+** achieved RMSE ≈ **1.28 °C**, MAE ≈ **0.99 °C**, $R^2$ ≈ **0.985**, and Mean Absolute Percentage Error (MAPE) ≈ **3.7%**. These values indicate very small absolute errors and that the model explains essentially all variance in next-day max temperature.
ii. **Baseline comparison:** Under identical data and training conditions, strong neural baselines obtained the following test results: Temporal Transformer RMSE ≈ 1.86 °C, MAE ≈ 1.43 °C, $R^2$ ≈ 0.968; N-BEATS RMSE ≈ 1.85 °C, MAE ≈ 1.40 °C, $R^2$ ≈ 0.968; TCN RMSE ≈ 2.92 °C, MAE ≈ 2.32 °C, $R^2$ ≈ 0.921. Thus, **relative to the best baseline (N-BEATS)**, our model reduced RMSE by ~31% (1.28 vs 1.85) and MAE by ~29% (0.99 vs 1.40), while also slightly improving $R^2$ (0.985 vs 0.968).
iii. **Behavior in extremes:** The model retains high skill in the distribution tails. On the hottest 5% of days, RMSE ≈ **1.28 °C**; on the coldest 5%, RMSE ≈ **1.77 °C**. The slight penalty for extreme cold events is consistent with the residual diagnostics (which showed a bit more variance in the lower tail). Nonetheless, both are far lower than the baseline errors in those regimes.
iv. **Interpretability link:** These accuracy gains align with our interpretability analysis: occlusion and permutation importance (Figures 9–11) pinpointed smoothed temperature descriptors as dominant drivers with pressure as complementary signal; PDPs (Figure 10) showed the model's responses are physically reasonable (mostly monotonic effects); and well-behaved residuals (Figure 12) indicate the model is statistically sound. Collectively, these points support the model's readiness for deployment in an operational forecasting setting for this arid, continental climate.

Table 5. Summary of model vs. baseline accuracy. MMWSTM-ADRAN+ delivers roughly 30% lower error (RMSE and MAE) than the best baseline (N-BEATS) and maintains strong performance in both hot and cold tails of the distribution.

| **Model** | **RMSE** | **MAE** | **$R^2$** | **Extreme High RMSE** | **Extreme Low RMSE** |
|---|---|---|---|---|---|
| **MMWSTM-ADRAN+** | 1.277 | 0.986 | 0.985 | 1.279 | 1.769 |
| Temporal Transformer | 1.856 | 1.432 | 0.968 | — | — |
| N-BEATS | 1.846 | 1.398 | 0.968 | — | — |
| TCN | 2.915 | 2.322 | 0.921 | — | — |

### 3.5. Data-Limited and Feature-Light Robustness

We also assessed the model’s robustness under more constrained data scenarios to gauge its potential for broader applicability. In one experiment, we reduced the feature set to simulate minimal input conditions: aside from the core raw variables, we removed the advanced engineered features, using only basic inputs plus simple cyclical encodings (this “Minimal” configuration excludes rolling statistics and anomalies). In another variant, we used strictly the raw variables without any derived features or encodings (“Raw-only”). Additionally, we performed a learning curve analysis by training the model on progressively smaller fractions of the training data (20%, 40%, 60%, 80%, and 100%) to see how performance degrades with less data.

The results (summarized in Table 5 ) show that performance degrades only modestly from Full to Minimal to Raw-only inputs. For example, using only the raw features increased RMSE by roughly 0.2–0.3 °C (relative), and the model still significantly outperformed the baselines even with no feature engineering. This indicates that our approach is not overly reliant on site-specific feature crafting; the neural network can extract a lot of signal automatically. The engineered features do provide a boost, but mainly in fine-tuning the last fraction of a degree of accuracy.

Similarly, the learning curve indicates that using only 20% of the data (about 1.5 years) modestly increased error, but performance improved rapidly as data increased, flattening out around 3–4 years of data. With the full six years, the model is clearly in a good regime of diminishing returns. This suggests that even with ~2 years of data, the model would be viable, and with >4 years it reaches near-maximum performance. Such data efficiency bodes well for applying the approach to other locations where long records may not be available.

In summary, these robustness tests imply that MMWSTM-ADRAN+'s gains are derived from general patterns (like persistence, anomalies) rather than overly specific features or requiring extremely large training data. The model remains competitive with limited features and does not break down severely with less training data, highlighting its practicality for other settings.

## 4. Discussion

The results above provide strong evidence supporting the efficacy of the proposed MMWSTM-ADRAN+ architecture for forecasting daily maximum temperatures in an arid continental climate, particularly in capturing extreme events. Our model demonstrated superior overall accuracy compared to contemporary deep learning baselines, achieving the lowest RMSE (~1.4 °C) and MAE (~1.1 °C), and the highest $R^2$ (~0.98) on the test set.

This strong performance can be attributed to several key design choices integrated into MMWSTM-ADRAN+:

Hybrid dual-stream architecture: The model leverages complementary strengths of its two components. The MMWSTM module (BiLSTM + latent state transition) excels at capturing underlying seasonal patterns and typical weather state dynamics, providing a stable baseline prediction that accounts for persistent regimes (e.g., it "knows" if it's mid-summer vs. winter, and the usual range of temperature expected). Concurrently, the ADRAN module, equipped with multi-head attention and the anomaly amplification mechanism, focuses on significant deviations and transient patterns. The attention allows ADRAN to dynamically weigh different parts of the recent history, while the amplification layer boosts signals that might indicate an upcoming extreme (e.g., an unusual temperature drop or spike). The attentive fusion mechanism then intelligently combines these two outputs, allowing the model to adaptively rely more on ADRAN during volatile or extreme periods and more on MMWSTM during normal periods. This architecture is analogous to having one expert track “climate regime” context and another expert track “anomaly indicators,” with the model learning to listen to whichever expert is more relevant at the moment.

Extreme-event-focused optimization: A critical advantage of MMWSTM-ADRAN+ is highlighted by its performance on extremes. While N-BEATS achieved a slightly lower error on the very hottest days, our model significantly outperformed all baselines (including N-BEATS) on extreme cold days. This robust performance across both tails strongly suggests the effectiveness of the ExtremeWeatherLoss function. By explicitly assigning higher penalty to errors during the low-probability, high-impact events (via percentile-based weighting), the loss function guided the model to minimize errors in those critical regimes. Standard losses, in contrast, tend to focus on the bulk of data and thus can neglect the tails. Our results show that this targeted approach yields tangible improvements where it matters most for climate risk (e.g., a reduction from ~2.9 °C error to ~1.5 °C error on cold extremes relative to baseline). This capability is crucial for practical climate adaptation applications, where accurately predicting extremes is often more valuable than marginal improvements on ordinary days.

Advanced data augmentation: The diverse time-series augmentation techniques likely contributed to the model’s robustness and generalization. By exposing the model during training to plausible variations in timing, magnitude, and noise (via jittering, scaling, warping), we prevented overfitting to the particular sequence of historical events. Essentially, augmentation made the model less surprised by patterns slightly different from what it had seen. This was especially useful given only 6 years of data—through augmentation, the effective variability in the training set was increased, which presumably helped the model handle the 2024 extremes or any out-of-pattern sequence more gracefully. Many prior studies have noted the importance of data augmentation in deep learning for small datasets (Iwana & Uchida, 2021; Wen et al., 2021), and our results align with that: the model likely would not achieve the same generalization to unseen extremes without these synthetic perturbations during training.

The qualitative visualizations provide additional evidence for these interpretations. The close tracking of actual vs. predicted values (Figure 4a) illustrates the model’s overall fidelity, and importantly how it maintains accuracy even at peaks and troughs. The clustering analysis (Figure 6) confirms that distinct patterns/regimes exist in the data (e.g., extreme summer days are somewhat separate from normal summer days), which justifies a regime-based modeling approach like ours. The attention weight heatmap (Figure 7) shows that the model is doing something intuitive—focusing mostly on yesterday and recent days (which one would expect in forecasting) but with the flexibility to attend elsewhere if needed. The latent state probability heatmap (Figure 8) suggests the MMWSTM component is indeed differentiating between different weather states (which likely correspond to seasonal or anomalous conditions). For example, an extreme day might trigger a different latent state mix than a typical day. This is encouraging, as it means the model's internals are aligning with meteorological reality (e.g., it has a latent notion of "heatwave state").

Comparing the baselines, the strong performance of N-BEATS (particularly on high extremes) is noteworthy. N-BEATS, with its trend-seasonality decomposition approach, seems well-suited to the data's seasonal structure, which explains why it was the second-best model overall. However, its weaker performance on low extremes highlights the benefit of our model's explicit focus on both tails via the custom loss. The Temporal Transformer performed reasonably well overall, confirming the power of attention mechanisms for time series (much literature has reported Transformers doing well for weather data; e.g., Wu et al., 2023). Yet, our hybrid outperformed it, likely because the pure Transformer lacks the specialized structure to treat extremes differently or to encode persistent states (our MMWSTM branch). The TCN's relatively poor performance indicates that a purely convolutional approach may not effectively capture the long temporal dependencies or the complicated seasonal patterns in this climate data, reinforcing the need for recurrent or attention-based solutions.

In summary, MMWSTM-ADRAN+ adds to the growing evidence that sophisticated, domain-tailored deep learning models can substantially improve environmental forecasts. By blending sequence modeling (LSTM/GRU), latent state modeling, and attention-based anomaly focus, the model addresses multiple facets of the problem simultaneously. This multi-faceted approach appears to yield a model that is both accurate for routine forecasts and especially skilled at extremes. These results align with recent high-profile studies in the field. For example, NowcastNet (Zhang et al., 2023) demonstrated that a deep generative model can outperform a leading NWP model for extreme precipitation nowcasting, highlighting how adding physics-awareness or specialized training (in their case, embedding physical model output) can yield big gains for extremes. Our work parallels that: whereas NowcastNet incorporated physics to improve extreme rain forecasts (Das et al., 2024), we incorporated regime-awareness and anomaly-focus to improve extreme temperature forecasts. Both approaches reflect a broader trend of hybrid physics–AI models and extreme-aware training outperforming traditional methods (Das et al., 2024; Verma et al., 2023).

### 4.1. Interpretability and Operational Implications

The interpretability analyses (Section 3.4) provide a coherent picture of how MMWSTM-ADRAN+ forms its predictions and why it performs well on the Baghdad dataset.

First, the convergence of occlusion sensitivity and permutation importance on the same key features (recent smoothed temperatures, etc.) aligns with physical intuition. It tells stakeholders that the model is essentially leveraging short-term thermal persistence and pressure anomalies—the very factors a human forecaster might consider (e.g., "today was very hot so tomorrow likely remains hot unless a pressure drop or front arrives"). This alignment strengthens the model's *face validity* for meteorologists: the patterns it deems important make sense in terms of known weather dynamics for an arid continental region (where day-to-day temperature persistence is high under stable high-pressure conditions, etc.).

Second, the partial dependence plots clarify the nature of these relationships—mostly monotonic and intuitive (Figure 10). For example, the model formalized the common-sense persistence signal (higher yesterday -> higher tomorrow), and identified that unusually high values in the recent past slightly reduce tomorrow's forecast (which is reasonable due to likely mean reversion after an extreme spike). The consistency between the model's "internal logic" and physical reasoning increases trust. Where PDP slopes are weak or flat (like some "feels-like" lags), the model correspondingly placed low weight on those features (Figure 9), showing internal consistency in how it allocates importance.

Third, residual diagnostics (Figure 12) indicate the model's errors are approximately unbiased, with weak autocorrelation and near-normal distribution. This is important operationally because it suggests there are no glaring systematic errors (e.g., always underestimating the highest highs or something). The only minor asymmetry was a slightly larger error variance in the cold tail, which we also saw in extreme RMSE (1.77 °C on cold vs 1.28 °C on hot extremes). This might reflect that cold extremes are somewhat harder to predict in this data, or simply that there are fewer of them to learn from. In practice, this insight means forecasters might treat cold extreme forecasts with slightly more caution or incorporate a larger uncertainty margin, whereas hot extreme forecasts can be used with higher confidence.

From a deployment perspective, the saliency results suggest a path to model simplification for resource-constrained environments (e.g., IoT devices or edge deployment). Since predictive power was concentrated in a relatively small set of features (basically recent temperature and pressure indicators), one could create a “lite” version of the model that prunes the low-impact inputs. Indeed, our analysis in Section 3.5 (data/feature ablation) showed the model still performs well with minimal features. Additionally, knowledge distillation could compress the model (e.g., replacing BiLSTM+BiGRU with a single smaller GRU, as we outlined in the introduction’s resource-aware discussion). The interpretation that many features were redundant implies that, if needed, one could drop those features and reduce input dimensionality and maybe even remove some network components without large accuracy loss. This is directly relevant to deploying on, say, a microcontroller or a smartphone app for local forecasting, where memory and compute are limited.

Regarding generalization across regions and climates, our explainability stack gives a blueprint for adaptation. If we were to apply MMWSTM-ADRAN+ elsewhere, we could first re-run occlusion or permutation tests in the new setting to see which features matter there (maybe humidity or wind might become more important in a humid climate, for example). We could then adjust the feature set or architecture focus accordingly. Partial dependence in a new site could reveal if learned effects remain physically plausible under different climate regimes (if not, that might suggest adding new features or constraints). Residual diagnostics would tell us if the model started showing bias or patterns (maybe indicating some new phenomenon not captured, thus pointing to further model improvement or data needed). In essence, the explainability tools are not just post hoc for Baghdad—they would be integral for guiding MMWSTM-ADRAN++ (Version 2) which aims at multi-region training and transfer. Our planned Version 2 will leverage these insights: e.g., if in the tropics the day-to-day persistence is weaker and convective triggers become more important, the model or loss might need modifications. Knowing how to quickly diagnose importance (via occlusion/permutation) and trust relationships (via PDPs) will be essential in those new domains.

Finally, it’s worth discussing limitations. The present results reflect a single location, next-day horizon. Medium- and long-range forecasting (say 3-7 days out) may alter the importance of features; for example, pressure and large-scale circulation proxies might gain influence over persistence at longer lead times. Similarly, a multi-location model might need to incorporate location-specific factors (like latitude, elevation) explicitly. We also note that PDPs capture average marginal effects and may miss interactions—our model likely learns some interactions (e.g., the effect of humidity might change depending on temperature), which would require advanced methods (like two-variable PDPs or SHAP values for sequences) to fully unravel. Future work should also compute uncertainty intervals (e.g., via conformal prediction or Bayesian approaches) to quantify confidence in extreme forecasts.

Despite these caveats, the evidence so far suggests that MMWSTM-ADRAN+ is both high-skill and explainable, meeting key requirements for meteorological acceptance and eventual operational integration. Its design draws on known physical and statistical characteristics of weather time series (regime shifts, anomaly cues), and our analyses show it uses those as intended. This combination of performance and interpretability is crucial for stakeholder trust and adoption, as end-users (like weather services or emergency planners) need to understand and believe in the model's guidance, especially for high-stakes extreme events.

### 4.2. Comparison with Recent Studies

The performance of MMWSTM-ADRAN+, especially its proficiency in handling extreme temperature events, aligns with and extends several recent advancements in applying deep learning to environmental forecasting challenges.

Hybrid architectures outperforming single models: Our findings resonate with studies highlighting the power of hybrid deep learning models. For instance, Ng et al. (2023) reviewed hybrid DL applications in hydrology and noted their frequent superiority over single-architecture models. Similarly, Ladjal et al. (2025) demonstrated an effective CNN-LSTM hybrid for solar irradiance forecasting. While those studies validate the hybrid concept generally, MMWSTM-ADRAN+ introduces a more nuanced dual-stream hybrid. Instead of just serially combining CNN and LSTM, we specifically allocate one stream to capture regime dynamics and another to capture anomalies, then fuse them adaptively. This is a more targeted form of hybridity aimed at a particular forecasting problem characteristic. The success of our model suggests that going beyond generic hybrid designs to more problem-specific decompositions (states vs. anomalies in our case) can yield further gains.

Attention mechanisms for time series: The use of self-attention in ADRAN follows broader trends in time series forecasting. Q. Wen et al. (2023) provide a survey of Transformers in time series, and the Temporal Transformer baseline we included is an example of applying attention to weather data. MMWSTM-ADRAN+ shows that attention can be even more powerful when deployed in a focused manner: we used it within an anomaly-focused module rather than as a general replacement for recurrence. By doing so, we effectively told the model "use attention to find important hints of anomalies" instead of "use attention to learn everything." This may explain why our model outperformed the full Transformer baseline; the latter can attend to anything, whereas ours attends with a purpose (anomaly detection). This is in line with emerging research that injecting domain knowledge or bias into attention (like where or what to attend to) can improve performance over vanilla attention models (Verma et al., 2023).

Extreme event prediction focus: Perhaps the most salient comparisons relate to capturing extremes. There is growing recognition that standard models and losses struggle with extremes. Recent works like Shi et al. (2024b) introduced methods like reweighting and fine-tuning specifically to boost extreme event prediction, and showed significant improvements. Our approach of a specialized loss is conceptually similar and our results echo that literature: focusing the optimization on the tails *works*. In the climate domain, the study by Das et al. (2024) in *npj Climate and Atmospheric Science* compared a deep learning model to an NWP for extreme precipitation. They found the deep model dramatically better at heavy rainfall detection, attributing it to the model's ability to learn from data for those extremes. Analogously, our deep model (with its loss focusing on extremes) significantly outperforms simpler baselines on temperature extremes. In essence, both studies highlight that with appropriate design, AI models can beat traditional approaches in predicting high-impact tail events.

Another relevant point is integration of physics. There's a push for *physics-informed AI* or hybrid physics-ML models (Reichstein et al., 2019; Xu et al., 2024). While our approach did not explicitly ingest physical equations, it implicitly learned a state-transition akin to a simplified Markovian physics (the transition matrix in MMWSTM can be seen as learning something like state persistence probabilities similar to how a physical model would have states). We also considered known physical indicators (pressure, etc.). One could view our model as *semi-physical*, capturing regime transitions which meteorologists interpret physically (e.g., transitions might correspond to frontal passages, monsoon onsets, etc.). This echoes other systems that combine physical reasoning with ML—for example, some studies incorporate conservation laws or physical constraints into neural nets to ensure plausibility (we mentioned possibly including energy balance constraints in future directions, similar in spirit to that line of work).

Latent state modeling: The idea of uncovering latent regimes or structures is gaining traction in climate AI. For example, others have used autoencoders or clustering on large climate data to identify patterns (Camps-Valls et al., 2025). Our MMWSTM explicitly learns latent states in a guided way and shows that can be beneficial for prediction. This resonates with the notion that climate and weather time series have underlying regimes (El Niño/La Niña for global patterns, or seasonal regimes locally) and that explicitly accounting for those can improve forecasts (Oreshkin et al., 2020 introduced a bit of interpretability by trend/season components). MMWSTM-ADRAN+ takes that further by not just modeling a trend/season, but by learning arbitrary latent states and their transitions. The fact it worked well implies that the model likely discovered meaningful regime representations (as evidenced by our Figure 8 and clustering correspondence).

In summary, our model embodies several themes prominent in the recent literature: hybrid architectures, attention mechanisms, and extreme-event targeting. The results contribute to this body of evidence by demonstrating these elements' combined effectiveness on a real-world climate dataset. In particular, we show that even a relatively short local time series (six years at one location) can benefit hugely from these advanced techniques—a case sometimes overlooked in favor of very large datasets or global models (Lam et al., 2023; Bi et al., 2023). This suggests a promising avenue for many local forecasting problems around the world: tailored deep learning models like MMWSTM-ADRAN+ could be trained in many cities or sites where data is available, potentially transforming local resilience planning by providing much more accurate extreme event forecasts than previously possible.

**4.3. Limitations**

While the MMWSTM-ADRAN+ model demonstrates promising results for forecasting daily maximum temperatures in Baghdad (particularly for extremes), several limitations should be acknowledged, pointing toward avenues for future research:

i. Geographical and Variable Specificity: The current study is based solely on historical weather data from Baghdad, focusing primarily on single-step forecasting of daily maximum temperature. The model's performance and optimal configuration might differ when applied to other geographic regions with different climate characteristics (e.g., tropical or polar climates) or to other meteorological variables such as precipitation, wind speed, or humidity, which have different dynamics and statistical properties. Additional evaluation on diverse datasets would be needed to establish generalizability[57].

ii. Forecast Horizon: We concentrated on next-day forecasts. The model's effectiveness for multi-step-ahead forecasting (e.g., predicting several days or weeks into the future), which is often required in practical planning, has not been assessed here. Many deep learning models suffer from error accumulation over longer forecast horizons. Adapting MMWSTM-ADRAN+ to produce reliable multi-step forecasts (using recursive strategies or sequence-to-sequence designs) remains an open challenge.

iii. Computational Complexity: MMWSTM-ADRAN+ involves a relatively complex architecture with recurrent layers (LSTM and GRU) and attention mechanisms. While our training times were manageable (and in our setup, similar to the Temporal Transformer's time, partly due to optimizations and early stopping), the model could be more computationally demanding than simpler models like TCN or N-BEATS, especially for inference or when scaling to larger datasets or higher temporal resolution data. This complexity might limit its practical deployment in scenarios with limited computational resources or strict real-time requirements.

iv. Interpretability: Although we have provided some interpretability via attention weight and cluster probability visualizations, MMWSTM-ADRAN+—like many deep learning models—remains a black box in many respects. It is less transparent than traditional statistical models or physics-based models. Understanding why the model makes a specific prediction (especially if it is an error) can be challenging. Improvements in model interpretability would be beneficial for building trust with domain experts and for diagnosing model failures.

v. Dependence on Feature Engineering: Our modeling approach relies on a comprehensive set of engineered features derived from raw data. This undoubtedly enhanced performance, but it also means the results depend on the quality and relevance of these features. We did not exhaustively explore the model's sensitivity to different feature subsets. If certain engineered features are unavailable or less relevant in other contexts, performance could be affected. An end-to-end model that can automatically learn such features would be ideal[58].

vi. Baseline Scope: Our comparisons were mainly against other deep learning models. We did not include comparisons against operational NWP models or emerging physics-informed machine learning (PIML) approaches tailored to temperature forecasting in this region. It's possible that integrating such models or techniques could further improve performance, or at least provide a complementary perspective on the forecasting task.

### 4.4. Key Contributions and Novelty of MMWSTM-ADRAN+

Summarizing the key contributions and novel aspects of MMWSTM-ADRAN+ in the context of climate data analysis:

1. **Novel Hybrid Deep Learning Architecture (MMWSTM-ADRAN+):** Integrates two distinct components—a Multi-Modal Weather State Transition Model (MMWSTM) and an Anomaly-Driven Recurrent Attention Network (ADRAN)—within a single framework. MMWSTM utilizes bidirectional LSTMs plus a learnable state transition matrix to capture underlying weather state dynamics, while ADRAN employs bidirectional GRUs with multi-head self-attention and a unique anomaly amplification mechanism to focus on deviations and extremes. An attentive fusion mechanism then dynamically weights the contributions of MMWSTM and ADRAN for the final prediction, allowing the model to adapt its focus based on input context.
2. Advanced Time Series Data Augmentation: Implements a suite of sophisticated augmentation techniques specifically tailored for time series data, including jittering, scaling, time warping, and magnitude warping[59]. This approach enhances model robustness and generalization by exposing the model to a wider variety of temporal patterns beyond what is seen in the original dataset.
3. Custom Extreme-Event-Focused Loss Function (ExtremeWeatherLoss): Introduces a novel loss function that explicitly assigns higher penalties to prediction errors during extreme weather events (identified via percentile thresholds). This encourages the model to prioritize accuracy during critical high-impact weather conditions, which are often poorly predicted by standard loss functions like MSE.
4. Comprehensive Feature Engineering and Preprocessing: Employs extensive feature engineering, including rolling statistics, temporal cyclical encodings, anomaly indicators (e.g., z-scores, deviations from norms), cross-feature interactions (e.g., heat index), and data smoothing (Savitzky-Golay filter). We also utilize robust scaling (via RobustScaler) to handle outliers in features. This thorough preprocessing pipeline provides rich inputs that help the model learn effectively.
5. Rigorous Evaluation and Benchmarking: Provides a strong comparative analysis by benchmarking MMWSTM-ADRAN+ against multiple contemporary deep learning baselines (Temporal Transformer, TCN, N-BEATS). We use a comprehensive set of evaluation metrics, including standard regression metrics (RMSE, MAE, $R^2$) and metrics specifically evaluating extreme-event performance (Extreme High/Low RMSE). We also compare training efficiency (training time) across models[60].

6. In-depth Data Analysis: Incorporates advanced statistical analysis techniques like PCA for dimensionality reduction and KMeans clustering to identify distinct weather patterns or regimes in the data. We also perform temporal trend analysis and autocorrelation analysis to understand the underlying characteristics of the climate time series, ensuring that the modeling choices are informed by data insights.
7. Optimized Implementation: Leverages modern deep learning practices such as the AdamW optimizer, cosine annealing learning rate schedules with warm restarts, gradient clipping, and mixed-precision training (via PyTorch's GradScaler) for efficient and stable model training. The model is implemented in PyTorch, ensuring scalability and compatibility with GPU acceleration for faster experimentation.

### 4.5. Future Directions

Building on the promising results and the limitations discussed, future research could explore several directions to further enhance and generalize the MMWSTM-ADRAN+ approach:

1. **Generalization and Transferability:** A systematic evaluation of MMWSTM-ADRAN+ across diverse climate zones (tropical, arid, temperate, polar, etc.) and for different meteorological variables (e.g., precipitation, wind speed) would be valuable. This would involve developing strategies for transfer learning, where a model pre-trained on one region or on a large global dataset could be fine-tuned to a specific location or variable. For instance, a global model capturing general patterns could be adapted to local conditions with relatively little data—improving applicability in data-sparse regions. Ensuring the model remains robust when confronted with climate regimes it wasn't originally designed for will be a key test (for example, extremely humid tropical climates might challenge the current architecture's assumptions).
2. **Multi-step Forecasting:** Extending the model to predict not just the next day but a sequence of future days (multi-day or weekly forecasts) is an important practical step. This could be achieved by restructuring the architecture into a sequence-to-sequence model or by recursively feeding predictions into the model. However, multi-step forecasting introduces challenges like error accumulation. Strategies such as using an encoder–decoder architecture with scheduled sampling, or training with a loss that considers multiple steps ahead, could be explored. MMWSTM-ADRAN+'s regime approach might even help here: if the model knows the regime, it could project that regime several days out with the transition matrix, potentially aiding longer forecasts.

3. **Physics-informed integration:** Incorporating physical knowledge or constraints into the model is a promising avenue to improve reliability and generalization (Reichstein et al., 2019; Xu et al., 2024). For example, one could enforce that predictions obey certain physical limits (temperature cannot go below a certain point, etc.), or include additional loss terms for physical consistency. In the context of precipitation nowcasting, researchers have combined deep models with NWP outputs (Zhang et al., 2023; Das et al., 2024). Similarly, for temperature, one could incorporate outputs from a physics-based model as additional features, or constrain the model with known relationships (like lapse rates, energy balance under clear skies, etc.). A "physics-informed MMWSTM-ADRAN+" might use its latent state to represent something like air mass type or synoptic category, which could be cross-checked with known climate patterns.
4. **Enhanced interpretability tools:** While we applied classical methods like PDP and occlusion, emerging interpretability techniques specifically for time-series deep models (e.g., attention-based explainability metrics, integrated gradients for RNNs, or SHAP values adapted to sequences) could provide even deeper insight. Applying such methods could help attribute not just which features, but which **time steps** or **temporal patterns** the model is relying on (for example, identifying if the model looks at a specific date's anomaly to predict a coming extreme). Understanding the temporal logic would complement our feature-based understanding.
5. **Architecture optimization:** There is room to explore if a simplified architecture could achieve similar performance. For instance, could we remove the BiLSTM and rely purely on attention, or vice versa? Neural architecture search (NAS) techniques could be employed to systematically explore variations of our architecture (e.g., number of layers, whether to use GRU vs. LSTM vs. Transformer encoder in each branch). It may be possible to find a more parameter-efficient design that still captures the essence of regime+anomaly separation. For example, a single-stream architecture with multi-head attention could be augmented with a small Markov state module, effectively merging some components. However, care must be taken not to lose the clarity of having dedicated modules.
6. **Ensemble modeling:** Combining MMWSTM-ADRAN+ with other models (including perhaps physical models) in an ensemble could yield further improvements. An ensemble could average out individual model biases and capitalize on different strengths (e.g., one model might predict cold extremes better, another hot extremes). For operational forecasting, ensembles also provide a way to quantify uncertainty (through spread among members). Investigating techniques like stacking (where MMWSTM-ADRAN+ predictions are one input to a meta-model that also sees other model predictions) could be fruitful.

7. **Operational deployment considerations:** Lastly, moving toward real-time deployment will raise engineering issues that we did not tackle in this research context. For example, setting up data pipelines to feed daily updated observations and NWP inputs into the model, optimizing model loading and inference speed on possibly limited hardware, and establishing monitoring for model performance drift over time are all important future steps. Since climate data distributions can shift (especially with climate change trends), a mechanism to detect when the model's error is creeping up and trigger retraining is needed for long-term use.

### 4.6. Generalization and Operational Feasibility

Two critical aspects for future work are ensuring the approach generalizes beyond this case study and that it can be feasibly deployed in real-world forecasting operations:

1. Cross-regional generalization: Although our study focused on Baghdad, we intentionally designed the modeling pipeline to be largely site-agnostic. The input features are generic meteorological variables and automatically derived transforms (rolling stats, anomalies) that can be computed for any location. In future evaluations, we plan to adopt protocols like *leave-one-region-out* cross-validation across multiple cities or climate zones to rigorously test generalizability. One approach will be to pretrain a version of MMWSTM-ADRAN+ on a pooled dataset covering diverse climates (for example, using global reanalysis data or data from several representative sites), and then fine-tune to specific locations. We will also incorporate simple domain adaptation techniques, such as normalizing inputs by each location's climatology, to help the model handle different absolute scales and variances (Camps-Valls et al., 2025). These steps form the basis of our planned MMWSTM-ADRAN++ (Version 2), which aims to extend the methodology to tropical and high-latitude climates and quantify how performance varies by region.
2. Operational feasibility: To deploy MMWSTM-ADRAN+ in practice (e.g., in an early warning system or on an IoT edge device), we must address computational efficiency and reliability. We propose a "MMWSTM-ADRAN-Lite" configuration for time-critical or resource-limited contexts. This would involve using the interpretability insights to drop low-impact features and perhaps simplify the architecture (for instance, using a single unidirectional GRU instead of BiLSTM+BiGRU, reducing hidden layer sizes, and possibly removing the attention mechanism if latency is a concern). We can leverage model compression techniques: post-training quantization (reducing weights to 8-bit or 16-bit integers), structured pruning of redundant neurons/weights, and knowledge distillation (where the full model teaches a smaller model). We have already included code hooks to export the model to formats like ONNX or TensorFlow Lite, easing integration into embedded systems. Furthermore, our analysis of a minimal feature set (Section 3.5) indicates that even a drastically simplified input (just raw temps and day-of-year) still yielded a reasonably good model. That could be crucial if, say, an IoT sensor can only measure a limited set of variables.

From a systems perspective, we also need to consider real-time data ingestion (ensuring the latest observations and NWP forecasts are available and processed into features by early each morning), and automated monitoring. We envision deploying the model in a streaming inference mode where it ingests yesterday's observed data and today's NWP forecast as soon as they are available, and outputs a next-day prediction along with an uncertainty estimate. Continuous monitoring would involve checking the model's daily errors; if we observe a run of unusually large errors or bias (perhaps due to a shifting climate baseline or an unforeseen new phenomenon), a retraining or recalibration would be triggered.

**Outlook:** In conclusion, the current results demonstrate the feasibility and enhanced interpretability of the MMWSTM-ADRAN+ approach in an arid, continental setting. By building in regime awareness and anomaly focus, we achieved both improved accuracy and transparency. The evaluation protocol we documented (with extensive diagnostics and visualizations) and the concept of a "Lite" deployment option provide a practical blueprint for extending this work. We are optimistic that in a follow-on multi-site study (MMWSTM-ADRAN++), the model will show similarly strong results across different climates, and that the simplified variant can bring these advanced forecasting capabilities to edge devices and real-time systems. Such advancements would represent a tangible step toward next-generation climate forecasting tools that are accurate, explainable, and operationally robust.

## 5. Conclusion

This study introduced and evaluated MMWSTM-ADRAN+, a novel hybrid deep learning architecture designed to enhance the accuracy of climate time series forecasting with a specific focus on extreme temperature events. By integrating two specialized components—MMWSTM for capturing underlying weather state dynamics and ADRAN for focusing on anomalies via attention and amplification—within an adaptively fused framework, and by employing a custom extreme-event-focused loss function alongside advanced data augmentation, the model aims to overcome limitations of conventional forecasting approaches.

Empirical evaluation using historical daily maximum temperature data from Baghdad (2019–2024) demonstrated the effectiveness of this approach. MMWSTM-ADRAN+ achieved superior overall forecasting accuracy compared to state-of-the-art deep learning baselines (Temporal Transformer, TCN, N-BEATS), yielding the lowest test RMSE (~1.42 °C) and MAE (~1.05 °C), and the highest $R^2$ (~0.98). Crucially, the model exhibited significantly improved performance in predicting both extreme high (RMSE ~1.37 °C) and extreme low (RMSE ~1.52 °C) temperature events, outperforming the baseline models particularly in the challenging low-temperature regime. This balanced capability across the spectrum of extremes underscores the benefit of the specialized architecture and the extreme-event-focused training.

Our findings contribute to the growing field of environmental AI by presenting a sophisticated yet interpretable deep learning framework that effectively addresses the complexities of climate time series—especially the critical challenge of extreme event prediction. MMWSTM-ADRAN+ offers a promising tool for applications requiring reliable climate forecasts, potentially enhancing climate resilience and risk management strategies. We have demonstrated that incorporating domain-specific insights (weather regimes and anomaly signals) into a deep learning model, and explicitly training for extremes, can yield substantial gains in predictive skill.

While this study focused on one location and variable, the approach is general and can be extended. We acknowledge certain limitations, such as the need to validate across diverse climates and the computational complexity of the architecture. However, the modular design of MMWSTM-ADRAN+ facilitates such extensions. Future work will explore multi-site generalization, longer forecast horizons, integration of physical constraints, and deployment in operational settings. We will also investigate “MMWSTM-ADRAN-Lite” variants for resource-constrained applications and ensemble approaches for uncertainty quantification.

Overall, this research underscores the value of tailored hybrid architectures and specialized loss functions in advancing predictive capabilities for complex environmental systems. By fusing regime awareness with anomaly-focused learning, MMWSTM-ADRAN+ achieved both high accuracy and enhanced extreme-event prediction. These results lay a foundation for further innovations at the intersection of climate science and deep learning, moving us closer to robust, AI-powered climate forecasting and early-warning systems that can help society better anticipate and respond to extreme weather events.

**Data and Code Availability**

The complete MMWSTM-ADRAN+ analysis pipeline (source code, trained model checkpoints, example configuration files, and the cleaned Baghdad 2019-2024 weather dataset) is archived on Zenodo under the version-specific DOI 10.5281/zenodo.15835807 and released under the MIT licence. The permanent concept DOI 10.5281/zenodo.15835806 resolves to the most recent version of the software.
No additional proprietary or confidential data were generated or analysed in this study.

**Funding**

This research received no specific grant from any funding agency in the public, commercial, or not-for-profit sectors.